%% file: 00-main.tex
\pdfoutput=1
\documentclass{article}

\PassOptionsToPackage{numbers, compress}{natbib}

\usepackage[final]{neurips_2022}

\usepackage[utf8]{inputenc} 
\usepackage[T1]{fontenc}    
\usepackage{hyperref}       
\usepackage{url}            
\usepackage{booktabs}       
\usepackage{amsfonts}       
\usepackage{nicefrac}       
\usepackage{microtype}      
\usepackage[dvipsnames]{xcolor}         
\usepackage{xspace}

\usepackage{bm}
\usepackage{bbm}
\usepackage{amsthm,amsmath,amssymb}
\usepackage{mathrsfs}
\usepackage{booktabs}
\usepackage{multirow}
\usepackage{bbding}
\usepackage{makecell}
\usepackage{enumitem}
\usepackage{caption}
\usepackage{subcaption}
\usepackage{graphicx}
\usepackage[detect-none]{siunitx}
\usepackage{wrapfig}
\usepackage{adjustbox}
\usepackage{color}
\usepackage{colortbl}

\definecolor{mygreen}{RGB}{46,139,87}
\definecolor{myblue}{RGB}{65,105,225}
\definecolor{lightgray}{RGB}{220,220,220}
\definecolor{darkgray}{RGB}{192,192,192}

\usepackage[capitalize]{cleveref}
\crefname{section}{Sec.}{Secs.}
\Crefname{section}{Section}{Sections}
\Crefname{table}{Table}{Tables}
\crefname{table}{Tab.}{Tabs.}

\newcommand{\rlip}{RLIP\xspace}
\newcommand{\ParSe}{ParSe\xspace}
\newcommand{\ParSeD}{ParSeD\xspace}
\newcommand{\rlipParSe}{RLIP-ParSe\xspace}
\newcommand{\rlipParSeD}{RLIP-ParSeD\xspace}
\newcommand{\spacing}{\\}

\title{RLIP: Relational Language-Image Pre-training \\ for Human-Object Interaction Detection}

\author{%
    Hangjie Yuan\textsuperscript{\rm 1}\thanks{Equal contribution. This work was done when Hangjie Yuan was an intern at DAMO Academy, Alibaba Group, supported by Alibaba Research Intern Program.} \hspace{.1in} Jianwen Jiang\textsuperscript{\rm 2}\footnotemark[1] \hspace{.1in} Samuel Albanie\textsuperscript{\rm 3} \\ 
    \textbf{Tao Feng}\textsuperscript{\rm 2} \hspace{.1in} \textbf{Ziyuan Huang}\textsuperscript{\rm 4} \hspace{.1in} \textbf{Dong Ni}\textsuperscript{\rm 1}\thanks{Corresponding author.} \hspace{.1in} \textbf{Mingqian Tang}\textsuperscript{\rm 2} \\ \\
    \textsuperscript{\rm 1}Zhejiang University \hspace{.2in} \textsuperscript{\rm 2}Alibaba Group \hspace{.2in} \textsuperscript{\rm 3}University of Cambridge \\
    \textsuperscript{\rm 4}National University of Singapore
    \\ \texttt{\{hj.yuan, dni\}@zju.edu.cn} \hspace{.2in} \texttt{sma71@cam.ac.uk} \hspace{.2in} \texttt{ziyuan.huang@u.nus.edu}\\
    \texttt{\{jianwen.jjw, shisi.ft, mingqian.tmq\}@alibaba-inc.com}
}

\begin{document}

\maketitle

\vspace{-0.3cm}

\begin{abstract}
The task of Human-Object Interaction (HOI) detection targets fine-grained visual parsing of humans interacting with their environment, enabling a broad range of applications.
Prior work has demonstrated the benefits of effective architecture design and integration of relevant cues for more accurate HOI detection.
However, the design of an appropriate pre-training strategy for this task remains underexplored by existing approaches. 
To address this gap, we propose \textit{Relational Language-Image Pre-training} (RLIP), a strategy for contrastive pre-training that leverages both entity and relation descriptions.
To make effective use of such pre-training, we make three technical contributions:
(1) a new \textbf{Par}allel entity detection and \textbf{Se}quential relation inference (ParSe) architecture that enables the use of both entity and relation descriptions during holistically optimized pre-training;
(2) a synthetic data generation framework, Label Sequence Extension, that expands the scale of language data available within each minibatch;
(3) mechanisms to account for ambiguity, Relation Quality Labels and Relation Pseudo-Labels, to mitigate the influence of ambiguous/noisy samples in the pre-training data.
Through extensive experiments, we demonstrate the benefits of these contributions, collectively termed RLIP-ParSe, for improved zero-shot, few-shot and fine-tuning HOI detection performance as well as increased robustness to learning from noisy annotations. 
Code will be available at \url{https://github.com/JacobYuan7/RLIP}.

\end{abstract}

\input{01-intro}
\input{02-related-work}
\input{03-method}
\input{04-experiments}

\input{05-conclusion}


\bibliographystyle{plain}
\bibliography{refs}

\input{08-appendix}

\end{document}

%% file: 01-intro.tex
\vspace{-0.2cm} 
\section{Introduction}
\vspace{-0.2cm} 


Driven by improvements in storage, sensors and networking technology, humanity is amassing vast archives of image and video data.
A significant fraction of this media is \textit{human-centric}---it is content focused on humans and their actions.  
The task of \textit{human-object interaction (HOI) detection}~\cite{chao2018learningtodetectHOI}
aims to provide a step towards fine-grained parsing of such content by detecting all possible triplets of the form <\textit{human, relation, object}> present in visual data. Robust HOI detection has myriad uses for image/video data analysis and represents essential functionality for visual and language applications such as image/video captioning~\cite{yao2018exploringVR,pan2020STgraph_for_videocap}, image retrieval~\cite{johnson2015imageRt}, image synthesis~\cite{johnson2018imagegeneration_SG} and video action understanding~\cite{ji2020actiongenome,yuan2021DIN}.

Given that sustained progress in object detection has yielded increasingly robust systems for detecting people and objects~\cite{ren2015faster,he2017mask,dai2021dynamic}, a key remaining challenge for HOI detection is to develop methods capable of generalising to the many possible pairs of interactions between these entities when provided with non-exhaustive training data.
To tackle this challenge, we draw inspiration from recent developments demonstrating that contrastive language-image pre-training can induce remarkable generalisation for zero-shot classification tasks~\cite{radford2021CLIP,Jia2021ScalingUV}.
These methods perform classification by casting it as a \textit{retrieval problem}, ensuring that the downstream task \textit{aligns closely} with the pre-training objective.
Recent work by Alayrac et al.~\cite{alayrac2022flamingo} hypothesises that it is this close alignment between the downstream and pre-training objectives that explains why contrastive methods have proven so effective for zero-shot classification.
In light of this hypothesis, in this work, we explore whether it is possible to achieve a similarly close alignment between the HOI detection task and its pre-training strategy.

While HOI detection has been widely studied~\cite{gupta2019nofrills,peyre2019HOIwithanalogy,xu2019HOIwithknowledge,liao2020ppdm,kim2020uniondet,kim2020actioncooccur, zhong2021GGNet,zou2021HOITransformer,kim2021hotr, tamura2021qpic,Yuan2022OCN,cong2022RelTR}, the topic of designing pre-training to reflect the final task objective remains under-explored.
Indeed, a widely adopted strategy~\cite{Yuan2022OCN,zhang2021CDN,tamura2021qpic,chen2021ASNet,zou2021HOITransformer} has been to employ object detection pre-training to initialise the parameters of the model responsible for both entity detection and relation inference.
However, while suitable for entity detection, such pre-training may be suboptimal for the detection of \textit{relations between entities} which often requires the model to take account of groups of entities with greater spatial context, rather than individual entities in isolation. 

To address this shortcoming of HOI detection, we propose \textbf{Relational Language-Image Pre-training} (RLIP) which tasks the model with establishing correspondences from both entities and relations to free-form text descriptions.
By doing so, RLIP endows the model with the ability to perform zero-shot HOI detection\footnote{\textit{Zero-shot} in this context refers to HOI detection without fine-tuning (following the terminology of~\cite{radford2021CLIP}), a formulation that assesses the generalization of a pre-training model to unseen distributions. This offers a practical alternative to the scenario (unseen combinations) considered in several prior HOI detection papers~\cite{hou2020VCL,hou2021FCL}.}.
Moreover, in contrast to previous pre-training schemes that are limited to predefined finite category lists, RLIP benefits from the rich descriptive nature of natural language supervision.

We encountered three barriers to a naive implementation of RLIP for existing methods:
(1) Recent end-to-end HOI detection architectures~\cite{tamura2021qpic,zhang2021CDN,zou2021HOITransformer,Yuan2022OCN} typically employ joint representations of (some subset of) \textit{subject, object} and \textit{relation} triplets.
As a consequence, it is difficult to leverage text descriptions for separate humans, objects and relations provided by existing datasets such as VG~\cite{krishna2017visualgenome}.
(2) Contrastive pre-training requires negative samples to train effectively, but it is unclear \textit{a priori} how such negatives should be constructed.
(3) Free-form text descriptions exhibit label noise and semantic ambiguity (since there can be many ways to describe the same concept in the absence of a canonical list of categories), rendering optimisation challenging. 

To overcome these barriers, we make several technical contributions in addition to the RLIP framework. 
First, to allow end-to-end contrastive pre-training with distinct descriptions of subsets, objects and relations, we propose the \textbf{Par}allel entity detection and \textbf{Se}quential relation inference (ParSe) architecture.
%
ParSe employs a DETR~\cite{carion2020DETR}-like design that allocates separate learnable query groups for subject and object representations, together with an additional set of conditional queries that encode relations.
While ParSe enables (and works best with) RLIP, we also find that it yields gains for traditional object detection pre-training schemes. 
To address the second barrier, we synthesise label sequences by extending
in-batch labels with out-of-batch sampling to ensure a plentiful supply of negatives---we term this Label Sequence Extension (LSE). 
For the third barrier, we exploit cross-modal cues to resolve label noise and relation ambiguity.
In particular, to mitigate label noise we use the quality of the visual entity detection phase~\cite{li2020GFLv1} to assign quality scores to relation-text correspondences, an approach we term Relational Quality Labels (RQL).
To mitigate relation ambiguity, we leverage similarities between labels to propagate relations via a pseudo-labeling scheme, which we term Relational Pseudo-Labels (RPL).

We demonstrate through experiments that relational pre-training outperforms traditional object detection pre-training schemes on comparable data.
We further find that the a zero-shot application of our combined approach, RLIP-ParSe, surpasses several existing fine-tuned methods.

%% file: 02-related-work.tex
\vspace{-0.2cm} 
\section{Related Work}
\vspace{-0.2cm} 
\paragraph{Human-object interaction detection.}
There is a rich body of work on HOI detection.
One theme has focused on the development of effective architectures for this task~\cite{liao2020ppdm,kim2020uniondet,tamura2021qpic,cong2022RelTR}.
A second theme has sought to leverage informative cues ranging from interaction points~\cite{liao2020ppdm,zhong2021GGNet}, interaction boxes~\cite{kim2020uniondet}, contextualized embeddings~\cite{tamura2021qpic,kim2021hotr,zou2021HOITransformer,kim2021hotr}, poses~\cite{gupta2019nofrills,li2019interactiveness} and statistical priors~\cite{kim2020actioncooccur,Yuan2022OCN} to external knowledge in the form of language embeddings \cite{xu2019HOIwithknowledge,peyre2019HOIwithanalogy,Yuan2022OCN}.
However, relational pre-training and open-vocabulary recognition remains underexplored.
The inter-pair transformations of IDN~\cite{li2020hoianalysis} and affordance transfer learning of ATL~\cite{hou2021ATL} can be interpreted as entity augmentations to train a stronger verb classifier, but these methods do not directly optimise for all the components of HOI detection.
Similar to ParSe, CDN~\cite{zhang2021CDN} explores disentangled embeddings.
However, it still couples the embedding of subjects and objects, rendering it suboptimal for RLIP.
The concurrent work, GEN-VLKT~\cite{GEN_VLKT} aims to derive knowledge from image-level language-image pre-training~\cite{radford2021CLIP}, while we aim to achieve aligned entity- and relation-level pre-training for HOI detection.

\textbf{Leveraging free-form text for visual pre-training.}
A series of recent papers have illustrated the significant value of employing free-form language to provide supervision for vision systems. 
CLIP~\cite{radford2021CLIP} and ALIGN~\cite{Jia2021ScalingUV} demonstrated striking improvements in zero-shot image classification ability through contrastive training of image-level representations.
Further work has sought to additionally leverage correspondences between objects/regions and text to learn flexible grounding models~\cite{kamath2021MDETR,li2021GLIP,zeng2021multi}.
We similarly seek to benefit from natural language supervision. 
However, to the best of our knowledge, we are the first to leverage correspondences between descriptions of relations and explicit pairings of \textit{subjects and objects} (rather than descriptions of images, objects or regions) as a pre-training signal.

%% file: 03-method.tex
\vspace{-0.2cm} 
\section{Methodology}  \label{sec:method}
\vspace{-0.2cm} 

\begin{figure*}[t]
\centering
\includegraphics[width=1\textwidth]{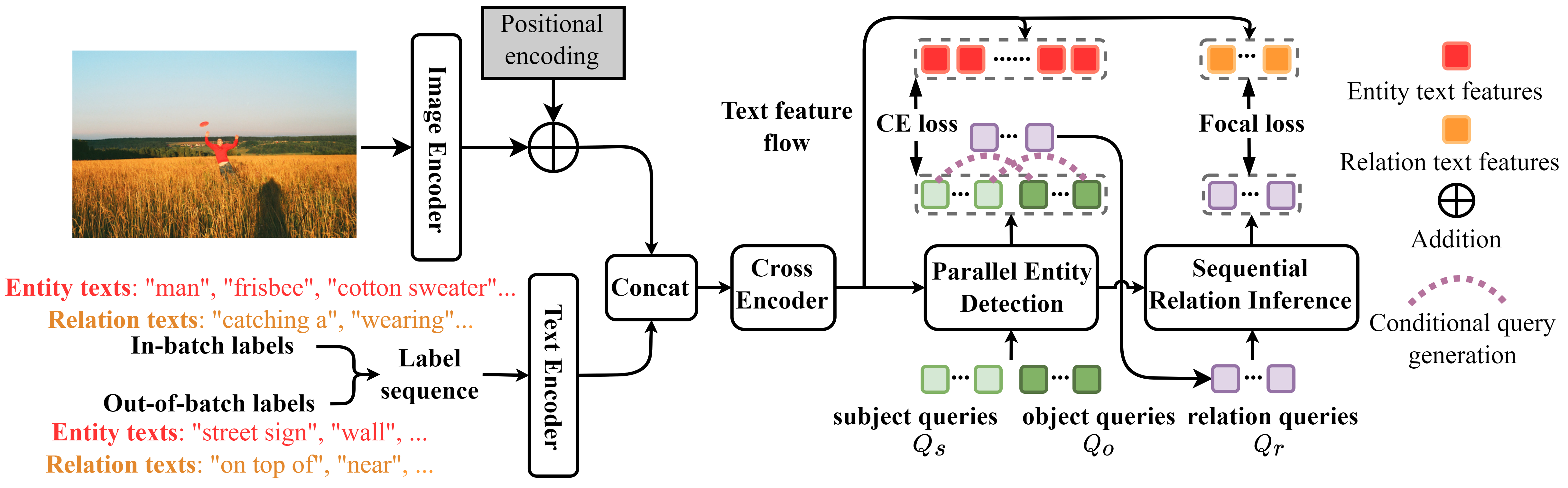}
\caption{\small An overview of our pre-training framework, RLIP-ParSe.
The \textit{Parallel Entity Detection} and \textit{Sequential Relation Inference} blocks represent independent DETR-style decoders responsible for entity and relation detection, respectively.
We omit the localisation loss for clarity.
See \cref{sec:method} for further details.}
\vspace{-0.4cm} 
\label{RLIP_pipeline}
\end{figure*}

In this section, we first present our triplet detection architecture, ParSe.
Second, we describe how ParSe is used to perform relational language-image pre-training (RLIP).
Finally, we introduce techniques to synthesise contrastive negatives and mitigate noise and ambiguities among labels.
The overall RLIP-ParSe framework is illustrated in~\cref{RLIP_pipeline}.

\vspace{-0.2cm} 
\subsection{ParSe for Triplet Detection}
\vspace{-0.2cm} 

\paragraph{Structure overview.}
The core idea underpinning the ParSe architecture is to allocate distinct representations of subjects, objects and relations in a holistically optimized model (rather than representing their combination, as commonly pursued in prior work~\cite{tamura2021qpic}). 
The motivation for doing so is two-fold:
(i) distinct representations enable the direct use of contrastive RLIP, since these representations can be put in correspondence with separate entity and relation annotations;
(ii) the separation of responsibilities allows for a more fine-grained control over the context available for each decision (a theme that has proven important for detection tasks~\cite{song2020Sibling_head}).
In particular, note that when detecting subjects and objects, local context is typically most useful.
However, when it comes to relations, detection will benefit not only from informative local cues, but also neighbouring context~\cite{yuan2021learningcontext} (for instance, it is useful to be aware of \textit{water} and \textit{hoses} when inferring the relation in the triplet \textit{<human, wash, car>}).
To instantiate this idea we follow ~\cite{zhang2021CDN} and implement triplet detection in a two-stage end-to-end manner.
Our probabilistic model factorises as follows:
\begin{equation}
\mathbb{P}(\bm{G}|\bm{Q}_{s}, \bm{Q}_{o}, \bm{C};\bm{\theta}_{Par}, \bm{\theta}_{Se}) 
= \mathbb{P}(\bm{B}_{s}, \bm{B}_{o}|\bm{Q}_{s}, \bm{Q}_{o}, \bm{C};\bm{\theta}_{Par})
\cdot
\mathbb{P}(\bm{R}|\bm{B}_{s}, \bm{B}_{o}, \bm{C};\bm{\theta}_{Se})
\label{eqn:factorisation}
\end{equation}
where $\bm{Q}_{s}, \bm{Q}_{o} \in \mathbb{R}^{N_Q \times D}$ define two sets of independent queries for $N_Q$ subjects and $N_Q$ objects;
$\bm{C}$ denotes features from the detection encoder;
$\bm{B}_{s}, \bm{B}_{o}, \bm{R}$ denote sets of detected subject boxes, object boxes and relations, respectively (these collectively comprise the detection results $\bm{G}$);
$\bm{\theta}_{Par}$ and $\bm{\theta}_{Se}$ represent learnable parameters from the entity detection decoder and the relation inferring decoder, respectively.
To construct ParSe, we design two components, \textit{Parallel Entity Detection} and \textit{Sequential Relation Inference}, to implement the second and third terms in \cref{eqn:factorisation}, respectively. These are described next.

\textbf{Parallel Entity Detection.}
Following the DETR family of architectures~\cite{carion2020DETR,zhu2020deformableDETR,liu2022DABDETR}, we first extract visual features using an image encoder, add positional encodings and then pass the result through a customized Transformer encoder according to the detector we adopt (we explore both DETR~\cite{carion2020DETR} and DDETR~\cite{zhu2020deformableDETR} variants) to obtain detection features $\bm{C}$.
Then, two sets of queries $\bm{Q}_{s}$ and $\bm{Q}_{o}$ are fed into the entity decoder to perform self-attention~\cite{AttentionAlluNeed},
cross-attention and feed-forward network (FFN) inference, obtaining $\tilde{\bm{Q}}_{s}, \tilde{\bm{Q}}_{o} \in \mathbb{R}^{N_Q \times D}$ which are used to predict box locations and classes.
\spacing 
\textbf{Sequential Relation Inference.}
To encode relations, we perform \textit{Sequential Relation Inference} as a sequential step after entity detection (similarly to \cite{zhang2021CDN}).
In the first stage, subjects and objects are detected via \textit{Parallel Entity Detection}.
In the second stage, we adopt a simple parameter-free matching scheme between subjects and objects to generate relation queries: matching by their indices.
Using this pairing scheme, we obtain relation queries via a conditional query generation function:
\begin{equation}
    \bm{Q}_r = F_{so}(\tilde{\bm{Q}}_{s}, \tilde{\bm{Q}}_{o})
\end{equation}
where for simplicity, we adopt addition as the query generation function.
Since we match by indices, $\bm{Q}_r \in \mathbb{R}^{N_Q \times D}$ contains $N_Q$ relation queries. 
$\bm{Q}_r$ is then fed into the second decoder to perform \textit{Sequential Relation Inference} via self-attention, 
cross-attention and FFN inference to obtain the corresponding relation features $\tilde{\bm{Q}}_r \in \mathbb{R}^{N_Q \times D}$ which are then used for relation classification.

\vspace{-0.2cm} 
\subsection{RLIP-ParSe for Relational Language-Image Pre-training}
\vspace{-0.2cm} 
For each iteration of pre-training, we construct a minibatch of images and their annotated relation triplets comprising all entities' locations, $N_{E}$ unique entity text labels as well as $N_{R}$ unique relation text labels. 
We describe how these are used for contrastive pre-training next.
\spacing 
\textbf{Formation of target label sequences.}
We construct targets from in-batch labels (free-form text descriptions forming subject, object, relation triplets).
In more detail, we first aggregate all entity labels within the batch and append to this sequence a \textit{no objects} label.
Next, we similarly aggregate all in-batch relation labels.
Then, all entity and relation labels are respectively fed into a text encoder (RoBERTa~\cite{liu2019RoBERTa} in our implementation) to extract label features denoted as $\bm{L}_{E}$ and $\bm{L}_{R}$, respectively.
Note that a free-form text label can have multiple tokens after tokenization---we use only the feature derived from the [CLS] token to represent the label. We concatenate the label feature sequence with features from the image encoder as shown in~\cref{RLIP_pipeline}. To fuse the concatenated features, we adopt a simple approach: applying a Transformer encoder \cite{li2021ALBEF,kamath2021MDETR,akbari2021VATT} to obtain fused label features $\tilde{\bm{L}}_{E} \in \mathbb{R}^{N_{E} \times D}$ and $\tilde{\bm{L}}_{R} \in \mathbb{R}^{N_{R} \times D}$.
\spacing
\textbf{Cross-modal alignment through classification.}
To implement RLIP, we task the model with establishing correspondences between entities/relations and their text descriptions using a classification objective, following~\cite{li2021GLIP}.
In particular, we align the $i$th relation $\tilde{\bm{Q}}_{r}(i) \in \mathbb{R}^{D}$ with relation its text via a Focal loss~\cite{lin2017focal}:
\begin{equation} \label{relation_focal}
    \bm{P}_{r}(i) = \tilde{\bm{Q}}_{r}(i) \tilde{\bm{L}}_{R}^{T} + \tilde{\bm{Q}}_{r}(i) \bm{W}_{b}^{T} + \bm{W}_{c}; \quad \mathcal{L}_{r}(j) = {\rm Focal({\rm sigmoid}}(\bm{P}_{r}(i, j)))
\end{equation}
where $\tilde{\bm{Q}}_{r}(i) \bm{W}_{b}^{T} + \bm{W}_{c}$ is the learnable bias term introduced in \cite{lin2017focal}; $\bm{W}_{b} \in \mathbb{R}^{N_{R} \times D}$ is a learnable linear projection and $\bm{W}_{c} \in \mathbb{R}^{N_{R}}$ is a constant vector filled with $-{\rm log}((1 - \pi) / \pi)$ with $\pi = .01$.
The Focal loss is defined via $\mathrm{Focal}(p) = - (1 - p)^\gamma \log (p)$ where $\gamma$ is set as a hyperparameter.
In the argument to this loss in \cref{relation_focal}, $j$ indexes along $\bm{P}_{r}(i) \in \mathbb{R}^{N_{R}}$.
To encourage matching of subjects and objects with their corresponding entity descriptions, an analogous objective to \cref{relation_focal} is used except that a softmax and a CE loss are applied and $\bm{W}_{c}$ is omitted (note that entities are uni-label and relations multi-label, as defined by the downstream task). The central benefit of the RLIP objectives defined above is that they bring the pre-training and downstream HOI detection losses into close alignment since the task of classifying entities and relations in the downstream task reflects the same matching task used in pre-training.
As a result, RLIP produces models that can perform HOI detection under \textit{zero-shot with no fine-tuning} (NF) evaluation protocols.
\spacing 
\textbf{Label Sequence Extension (LSE).}
Within a given batch, the number of negative samples available for matching is limited.
However, provision of plentiful negatives has been widely shown to improve contrastive learning~\cite{chen2020simCLR,feng2021LOOKNN,Islam2021study_on_transferability,wang2020uniformity}.
To this end, we propose Label Sequence Extension as a mechanism to leverage out-of-batch text descriptions.
Concretely, we sample additional text descriptions with a ratio of two thirds entity labels and one third relation labels.
To ensure computational tractability in the presence of the quadratic complexity of Transformer, we limit the label sequence to a predefined length $N_L$.
We experiment with two sampling strategies:
(i) \textit{Uniform sampling} that draws among candidate labels with equal probability;
(ii) \textit{Frequency-based sampling} that samples according to the label frequency in the training set.

\vspace{-0.2cm} 
\subsection{Addressing Relational Semantic Ambiguity}
\vspace{-0.2cm} 
Datasets with crowd-sourced language annotations~\cite{krishna2017visualgenome,kuznetsova2020open_image} exhibit significant label noise and ambiguity.
First, the descriptions themselves may be noisy (inaccurate), particularly when the underlying image is challenging to interpret.
A second challenge for traditional training schemes is that similar relations can be described differently, thanks to synonyms.
For example, the \textit{stand near} relation may be annotated ``stand near'', ``stand next to'', ``stand by'', \textit{etc.}.
These forms of \textit{semantic ambiguity} make supervised cross-modal pre-training (which relies on access to consistent labels) challenging.
To mitigate this issue, we focus on two aspects of the pre-training input data:
(i) the quality of the relation text labels;
(ii) the presence of semantically-similar labels in sampled label sequences.
\spacing
\textbf{Relational Quality Labels (RQL).}
To tackle the first challenge, we propose a label smoothing~\cite{lukasik2020LSR_mitigate} approach. 
The key idea is that we expect the difficulty of subject and object detection for a particular instance to correlate with the confidence of the annotated relation.
We therefore propose to estimate annotation quality from the quality of the entity detection stage.
Drawing inspiration from the generalised focal loss~\cite{li2020GFLv1}, we instantiate this idea by assessing the quality of the $i$th subject and object detection after bipartite matching~\cite{tamura2021qpic} as
\begin{equation}
    \bm{e}(i) = {\rm min}({\rm GIoU}_{\numrange{0}{1}}(\bm{B}_{s}(i), \hat{\bm{B}}_{s}(i)), {\rm GIoU}_{\numrange{0}{1}}(\bm{B}_{o}(i), \hat{\bm{B}}_{o}(i)))
\end{equation}
where ${\rm GIoU}_{\numrange{0}{1}}$ denotes generalized IoU from \cite{rezatofighi2019GIoU} together with a linear scaling function to scale the ${\rm GIoU}$ value to the range of 0 to 1 and $\,\hat{}\,$ denotes ground-truth annotation.
The resulting value $\bm{e}(i)$ is then employed to calibrate the relation label confidence via multiplication: $\Tilde{\bm{R}}(i) = \bm{e}(i) \hat{\bm{R}}(i)$.
\textbf{Relational Pseudo-Labels (RPL).}
To address the second issue, we propose a pseudo-labelling strategy~\cite{yarowsky1995unsupervised} to account for synonyms in the extended sequence.
We exploit the fact that text embeddings with high semantic similarity will lie close together, as measured by an appropriate distance function $M(\cdot ,\cdot)$.
We define the distance between the $i$th annotated relation label $\hat{\bm{R}}(i) = \{0, 1\}^{N_{R}}$ and the $j$th relation text feature $\tilde{\bm{L}}_{R}(j) \in \mathbb{R}^{D}$ from the extended sequence as
\begin{equation}
    M(\hat{\bm{R}}(i), \tilde{\bm{L}}_{R}(j)) = \sum\nolimits_{k = 1}^{N_{R}}{\hat{\bm{R}}(i, k)} \cdot m (\tilde{\bm{L}}_{R}(k), \tilde{\bm{L}}_{R}(j))
\end{equation}
where $m(\cdot, \cdot)$ denotes Euclidean distance.
Given the $i$th relation label, we apply a scaling function to $M(i, j)$ via $\bar{M}(i, j) = \frac{{\rm max}_{k}(M(i,k)) - M(i,j)}{{\rm max}_{k}(M(i,k))}$
where we have abbreviated $M(\hat{\bm{R}}(i), \tilde{\bm{L}}_{R}(j))$ as $M(i, j)$ for clarity.
Next, we use a global threshold $\eta$ to select label texts with high similarities: we set the $j$th label in the $i$th relation labels as $\bar{M}(i, j)$ if $\bar{M}(i, j) > \eta$.
Note that when applying either RQL or RPL, the ground truth labels are continuous (rather than discrete).
We therefore employ the Quality Focal Loss~\cite{li2020GFLv1} (rather than the standard Focal Loss~\cite{lin2017focal}) as our objective function.

\vspace{-0.2cm} 
\subsection{Pre-training, Fine-tuning and Inference}
\vspace{-0.2cm} 
By design, our pre-training (RLIP) and fine-tuning phases follow a similar process. 
For a given batch of images with corresponding annotations, we aggregate the results from \textit{Parallel Entity Detection} and \textit{Sequential Relation Inference} to form $N_Q$ triplets per image.
During pre-training and fine-tuning, we employ bipartite matching similarly to prior work~\cite{tamura2021qpic,zou2021HOITransformer,chen2021qahoi,zhang2021CDN,Yuan2022OCN}, following in particular the matching cost proposed in~\cite{tamura2021qpic}.
The overall loss is then constructed as follows:
\begin{equation}
    \mathcal{L} = \lambda_{1} \mathcal{L}_{l1} + \lambda_{2} \mathcal{L}_{GIoU} + \lambda_{3} (\mathcal{L}_{s} + \mathcal{L}_{o}) + \lambda_{4} \mathcal{L}_{r}
\end{equation}
where $\mathcal{L}_{l1}, \mathcal{L}_{GIoU}, \mathcal{L}_{s}, \mathcal{L}_{o}, \mathcal{L}_{r}$ denote the $\ell_1$ loss for box regression, GIoU loss~\cite{rezatofighi2019GIoU}, CE loss for subject and object classes, and Focal loss for relations (or Quality Focal loss~\cite{li2020GFLv1} when applying RQL or RPL), respectively.
The $\lambda$ terms are fixed weights to balance multi-task training following \cite{tamura2021qpic}, with $\lambda_{1}=2.5, \lambda_{2}=1, \lambda_{3}=1, \lambda_{4}=1$.
During pre-training, the label sequence is constructed from both in-batch and out-of-batch labels. 
During fine-tuning, we use all text labels contained in the dataset to form the label sequence (unlike pre-training, these labels fall within a pre-defined category list of limited size). 
Note that we follow~\cite{tamura2021qpic,zhang2021CDN} and exclude $\mathcal{L}_{s}$ during fine-tuning since HOI detection detects only humans as subjects. 
During inference, the confidence score for an object is simply the top-1 score from the softmax distribution over objects, and the relation score is obtained by multiplying the original score from the ${\rm sigmoid}$ function and the object score.
We rank relation scores and filter out the top-$K$ within those correctly localised triplets (IoU > 0.5) for evaluation.
$K$ is set to 100 by default following~\cite{tamura2021qpic,zhang2021CDN,liao2020ppdm,Yuan2022OCN}.

%% file: 04-experiments.tex
\vspace{-0.2cm} 
\section{Experiments} \label{sec:expriments}
\vspace{-0.2cm} 
\textbf{Datasets.}
We use the Visual Genome (VG)~\cite{krishna2017visualgenome} dataset for RLIP.
This dataset contains 108,077 images annotated with free-form text for a wide array of objects and relations (100,298 object annotations and 36,515 relation annotations).
The dataset is pre-processed prior to use (see supplementary material for details).
For downstream tasks, we conduct experiments on HICO-DET~\cite{chao2015hico} and V-COCO~\cite{gupta2015VisualSemanticRole}.
HICO-DET contains 37,536 training images and 9,515 testing images, annotated with 600 HOI triplets derived from combinations of 117 verbs and 80 objects.
We evaluate under the \textbf{Default} setting.
V-COCO comprises 2,533 training images, 2,876 validation images and 4,946 testing images annotated with 24 interactions and 80 objects.
Results are assessed under two scenarios denoted as ${\rm AP}_{role}^{\# 1}$ and ${\rm AP}_{role}^{\# 2}$ as defined by the official evaluation code~\cite{gupta2015VisualSemanticRole}. 
Note that all object classes in HICO-DET and V-COCO are identical to COCO.
\spacing 

\textbf{Implementation details.}
The basic architecture of the encoder and decoder are based on DETR~\cite{carion2020DETR} for \textbf{ParSe} and DDETR~\cite{zhu2020deformableDETR} for an additional variant named \textbf{ParSeD}.
A detailed architecture description of RLIP-ParSeD (which uses an additional transformer for cross-modal fusion) is provided in the supplementary.
For \textit{Parallel Entity Detection} and \textit{Sequential Relation Inference}, 3 decoding layers are used.
The number of queries $N_Q$ is set to 100 during pre-training and 64 during fine-tuning (following \cite{zhang2021CDN}). $\gamma$ in the Focal loss is set to 2 following~\cite{tamura2021qpic,zhang2021CDN}. 
$N_{L}$ in LSE is set to 500 to ensure computational tractability.
$\eta$ in RPL is set to 0.3.
For pre-training and fine-tuning, the initial learning rate (LR) of the image and text encoders is set to 1e-5, while all other modules are set to 1e-4.
For RLIP-ParSeD and ParSeD (object detection and relation detection pre-training), we pre-train model on VG for 50 epochs and drop LR by a factor of 10 at epoch 40. 
For ParSeD and RLIP-ParSeD, We fine-tune for 60 epochs and drop LR at epoch 40 by a factor of 10.
For ParSe and RLIP-ParSe, We fine-tune for 90 epochs and drop LR at epoch 60 by a factor of 10.
The pre-training and fine-tuning strategy follow above descriptions unless stated otherwise.
Experiments are conducted on 8 Tesla V100 GPU cards with a minibatch size of 32.

\textbf{Experimental protocols.} 
To assess performance under \textit{fine-tuning} and \textit{zero-shot with no fine-tuning} (NF) scenarios, we evaluate on HICO-DET across three HOI sets under the mAP metric:
\textit{Rare} (HOIs with training samples less than 10, of which there are 138),
\textit{Non-Rare} (HOIs with samples equal to or more than 10, of which there are 462)
and
\textit{Full} (all HOIs, of which there are 600).

To evaluate performance under the zero-shot formulation considered by~\cite{hou2021ATL,hou2021FCL,hou2020VCL}, we report results on unseen combinations (UC).
In particular, we report results under two settings:
\textit{UC with rare-first} (UC-RF) selection and \textit{UC with non-rare first} selection (UC-NF), both of which are assessed across three subsets: Unseen (120 HOIs), Seen (480 HOIs) and Full (600 HOIs). 

To evaluate few-shot transfer performance, we follow~\cite{kamath2021MDETR} and sample subsets of training annotations from HICO-DET. 
In detail, we sample $1\%$ and $10\%$ of the total annotations available among the HICO-DET training data, ensuring that all objects and verbs (but not all combinations) exist in the selected annotations.
Similarly to the fine-tuning protocol, we evaluate on the \textit{Rare}, \textit{Non-Rare} and \textit{Full} sets.

To assess the robustness of RLIP and its sensitivity to noise in the relation labels, we follow~\cite{Li2020DivideMix,Wu_2021_ICCV} and artificially inject noise into the relation labels by randomly flipping a fixed ratio of verbs in HOI triplets across the training set.

\begin{table}[t]
  \small
  \footnotesize
  \setlength{\tabcolsep}{2pt}
  \centering
  \caption{\small Comparisons with previous fine-tuned methods on \textcolor{myblue}{HICO-DET} and \textcolor{mygreen}{V-COCO} (column \textcolor{mygreen}{${\rm AP}_{role}^{\# 1}$} and \textcolor{mygreen}{${\rm AP}_{role}^{\# 2}$}). PT, PTP, OD, RD and MD abbreviate Pre-Training, Pre-Training Paradigm, Object Detection, Relation Detection and Modulated Detection. FRCNN, R, HG and Swin-T denote Faster R-CNN~\cite{ren2015FRCNN}, ResNet~\cite{he2016resnet}, Hourglass~\cite{newell2016hourglass} and Swin-Tiny~\cite{liu2021swin}. \textsuperscript{*} denotes RLIP is performed on VG, initialized with parameters from COCO object detection. \textsuperscript{\dag} denotes fine-tuning for 150 epochs following QAHOI~\cite{chen2021qahoi}.}
    \begin{tabular}{cccccccccc}
    \toprule
    \textbf{PT Data} & \textbf{PTP} & \textbf{Method} & \textbf{Detector} & \textbf{Backbone} & \textcolor{myblue}{\textbf{Rare}} & \textcolor{myblue}{\textbf{Non-Rare}} & \textcolor{myblue}{\textbf{Full}} & \textcolor{mygreen}{${\rm AP}_{role}^{\# 1}$} & \textcolor{mygreen}{${\rm AP}_{role}^{\# 2}$} \\
    \midrule
    \multirow{2}[1]{*}{-} & - & QAHOI\textsuperscript{\dag}~\cite{chen2021qahoi} & DDETR & Swin-T & 22.44 & 30.27 & 28.47 & - & - \\
          & -      & ParSeD\textsuperscript{\dag} & DDETR & Swin-T & \textbf{25.76} & \textbf{31.84} & \textbf{30.44} & - & -  \\
    \midrule
    \multirow{15}[1]{*}{COCO} & \multirow{15}[1]{*}{OD} & InteractNet~\cite{kaiming18DetectHOI} & FRCNN & R50-FPN & 7.16  & 10.77  & 9.94 & 40.0 & - \\
          &       & GPNN~\cite{Qi2018GPNN} & FRCNN & R152-DCN & 9.34  & 14.23  & 13.11 & 44.0 & - \\
          &       & iCAN~\cite{gao2018ican}  & FRCNN & ResNet-50 & 10.45  & 16.15  & 14.84 & 45.3 & 52.4 \\
          &       & UnionDet~\cite{kim2020uniondet} & RetinaNet & R50-FPN & 11.72  & 19.33  & 17.58 & 47.5 & 56.2 \\
          &       & PPDM~\cite{liao2020ppdm} & CenterNet & HG104 & 13.97 & 24.32  & 21.94 & - & - \\
          &       & HOTR~\cite{kim2021hotr} & DETR & ResNet-50 & 17.34 & 27.42  & 25.10 & 55.2 & 64.4\\
          &       & HOITransformer~\cite{zou2021HOITransformer} & DETR & ResNet-50 & 16.91 & 25.41  & 23.46 & 52.9 & - \\
          &       & QPIC~\cite{tamura2021qpic} & DETR & ResNet-50 & 21.85 & 31.23  & 29.07 & 58.8 & 61.0 \\
          &       & OCN~\cite{Yuan2022OCN}   & DETR  & ResNet-50 & 25.56  & 32.51  & 30.91 & 64.2 & 66.3 \\
          &       & CDN~\cite{zhang2021CDN}   & DETR  & ResNet-50 & 27.39 & 32.64  & 31.44 & 61.7 & 63.8 \\
          &       & QAHOI~\cite{chen2021qahoi} & DDETR & ResNet-50 & 18.06  & 28.61  & 26.18 & - & - \\
          &       & ParSeD & DDETR & ResNet-50 & 22.23  & 31.17  & 29.12 & 61.8 & 64.0  \\
          &       & ParSe & DETR  & ResNet-50 & 26.36  & 33.41 & 31.79 & 62.5 & 64.8 \\
          &       & ParSe & DETR  & ResNet-101 & \textbf{28.59} & \textbf{34.01} & \textbf{32.76} & \textbf{64.4} & \textbf{66.5} \\
    \midrule
    \multirow{3}[1]{*}{VG} & OD & ParSeD & DDETR & ResNet-50 & 19.59  & 25.03  & 23.78 & 41.4 & 43.0\\
          & RD  & ParSeD & DDETR & ResNet-50 & 21.36 & 29.27 & 27.45 & 51.5 &  53.2 \\
          & RLIP  & RLIP-ParSeD & DDETR & ResNet-50 & \textbf{24.45} & \textbf{30.63} & \textbf{29.21} & \textbf{53.1} & \textbf{55.0} \\
    \midrule
    \multirow{2}[1]{*}{COCO+VG} & RLIP\textsuperscript{*} & RLIP-ParSeD & DDETR & ResNet-50 & 24.67 & 32.50 & 30.70 & 61.7 & 63.8 \\
          & RLIP\textsuperscript{*}  & RLIP-ParSe & DETR & ResNet-50 & \textbf{26.85} & \textbf{34.63} & \textbf{32.84} & \textbf{61.9} & \textbf{64.2} \\
    \midrule
    GoldG+ &  MD & MDETR-ParSe~\cite{kamath2021MDETR} & DETR  & ResNet-101 & 22.91  & 31.07  & 29.19 & 53.6 & 56.0  \\
    \bottomrule
    \end{tabular}%
    \vspace{-0.4cm} 
  \label{SOTA_HICO}%
\end{table}%

\vspace{-0.2cm} 
\subsection{Results and Analysis}
\vspace{-0.2cm} 

\paragraph{Comparing object detection and relation detection pre-training with RLIP.} An assessment of the benefits of object detection pre-training using the COCO dataset may offer a somewhat optimistic evaluation of this approach, since COCO shares identical object classes with the downstream HOI detection evaluation datasets HICO-DET and V-COCO (and in the latter case shares training images).
If we control for this effect by performing object detection pre-training on VG rather than COCO, we observe a significant drop in performance for our ParSeD baseline (from 29.12 to 23.78 across the \textit{Full} set on HICO-DET and from 61.8 to 41.4 on V-COCO for ${\rm AP}_{role}^{\# 1}$, as shown in \cref{SOTA_HICO}).
However, since COCO lacks relation annotations, we investigate the benefits of RLIP on VG. 
We observe that RLIP outperforms vanilla object detection pre-training by a wide margin (boosting performance from 23.78 to 29.21 across the \textit{Full} set on HICO-DET and from 41.4 to 53.1 on V-COCO for ${\rm AP}_{role}^{\# 1}$), demonstrating the value of incorporating relations as a pre-training cue.
Another way to pre-train with relations is to perform relation detection which is still inferior to RLIP (27.45 < 29.21), demonstrating the importance of relational language-image pre-training.

\paragraph{Leveraging off-the-shelf object detection data without relations.}
As shown in the previous experiment, while VG provides relation annotations, it provides a much weaker basis for object detection pre-training than COCO.
More broadly, we may expect that object annotations are likely to be more readily available (and greater in scale) than relation annotations.
To mitigate this, a simple solution is to simply load object detection parameters pre-trained from an object-annotated dataset (like COCO), to complement the abilities of RLIP.
\cref{SOTA_HICO} indicates that RLIP-ParSeD indeed benefits from this approach, surpassing both object detection pre-training and RLIP (29.12, 29.21 $\rightarrow$ 30.70 on the \textit{Full} set).
We pre-trained on DETR for 150 epochs, outperforming an expert object detection pre-training (31.79 $\rightarrow$ 32.84).
On V-COCO, there is a degradation of performance (62.5 $\rightarrow$ 61.9) which we believe may be caused by the reduced domain alignment (i.e. common training images) relative to COCO object-detection pre-training~\cite{tamura2021qpic}.

\paragraph{Comparing cross-modal regional alignment pre-training with \rlip.}
We next compare to the use of language-region alignment pre-training introduced by MDETR~\cite{kamath2021MDETR}, which employed the GoldG+ dataset (this comprises VG, COCO and Flickr30k~\cite{plummer2015flickr30k} together with the corresponding annotations for referring expressions, VG regions, Flickr entities, and GQA~\cite{hudson2019GQA}).
For comparison, we initialise \rlipParSe with MDETR's parameters and then fine-tune on HICO-DET. The results are reported in \cref{SOTA_HICO} as MDETR-\ParSe.
Although MDETR makes use of a heavier backbone (ResNet-101) and additional pre-training data, \rlipParSe nevertheless surpasses this baseline with a lighter ResNet-50 backbone, demonstrating the effectiveness of RLIP for this task.

\paragraph{Zero-shot HOI detection.}
To assess performance under the zero-shot NF protocol, we compare with other methods using \rlipParSe (initialised with COCO parameters followed by \rlip on VG) and ParSe (initialised with COCO parameters).
Note that during RLIP, we match subjects against a diverse collection of categories.
However HOI detection only needs to detect \textit{person} as subjects.
Consequently, for zero-shot NF inference, we filter out subjects that are not classified as \textit{person}.
We report results in~\cref{tab:NF-protocol}, where we observe that RLIP outperforms several fully fine-tuned methods.
We find, as expected, that regional language-image pre-training methods like MDETR fail under an NF evaluation, since its pre-training lacks the notion of explicit relations.
Under UC-RF and UC-NF protocols (fine-tuning for 40 epochs under UC-NF to avoid over-fitting), \rlipParSe outperforms previous methods and ParSe by performing RLIP on VG.
\textbf{Few-shot transfer on HICO-DET.}
To evaluate few-show transfer, we fine-tune \ParSeD for 60 epochs as above, while RLIP-ParSeD is fine-tuned for 10 epochs to avoid over-fitting.
The results are shown in~\cref{few_shot_transfer_RLIP_ParSeD}.
We observe that RLIP significantly benefits few-shot fine-tuning relative to object detection pre-training and relation detection pre-training, especially when data is scarce. 

\begin{minipage}[!t]{\textwidth}
\begin{minipage}[c]{0.48\textwidth}
    \small
    \setlength{\tabcolsep}{1pt}
    \centering
    \makeatletter\def\@captype{table}\makeatother\caption{\small Results under zero-shot settings on HICO-DET. NR denotes Non-Rare.}
    \begin{tabular}{cccccc}
    \toprule
    \textbf{Zero-shot} & \textbf{Method} & \textbf{Rare} & \textbf{NR} & \textbf{Full} \\ 
    \midrule
    \multirow{2}[1]{*}{NF} & MDETR-ParSe~\cite{kamath2021MDETR} & 0.00  & 0.00  & 0.00  \\ 
          & RLIP-ParSe & \textbf{15.08} & \textbf{15.50} & \textbf{15.40} \\
    \toprule
    \textbf{Zero-shot} & \textbf{Method} & \textbf{Unseen} & \textbf{Seen} & \textbf{Full} \\ 
    \midrule
    \multirow{5}[2]{*}{UC-RF} & VCL~\cite{hou2020VCL}  & 10.06  & 24.28  & 21.43  \\
          & ATL~\cite{hou2021ATL}  & 9.18  & 24.67  & 21.57  \\
          & FCL~\cite{hou2021FCL}  & 13.16  & 24.23  & 22.01  \\
          & ParSe & 18.53 & 32.21 & 29.06 \\
          & RLIP-ParSe & \textbf{19.19} & \textbf{33.35} & \textbf{30.52} \\
    \midrule
    \multirow{5}[2]{*}{UC-NF} & VCL~\cite{hou2020VCL}  & 16.22  & 18.52  & 18.06  \\
          & ATL~\cite{hou2021ATL}  & 18.25  & 18.78  & 18.67  \\
          & FCL~\cite{hou2021FCL}  & 18.66  & 19.55  & 19.37  \\
          & ParSe & 19.65 & 24.50 & 23.38 \\
          & RLIP-ParSe & \textbf{20.27} & \textbf{27.67} & \textbf{26.19} \\
    \bottomrule
    \end{tabular}%
    \label{tab:NF-protocol}%
\end{minipage}
\begin{minipage}[c]{0.5\textwidth}
    \small
    \setlength{\tabcolsep}{1pt}
    \centering
    \makeatletter\def\@captype{table}\makeatother\caption{\small Few-shot transfer on HICO-DET. OD, RD denote object detection pre-training and relation detection pre-training.}
    \begin{tabular}{cccccc}
    \toprule
    \textbf{Method} & \textbf{Data} & \textbf{Epochs} & \textbf{Rare} & \textbf{Non-Rare} & \textbf{Full} \\
    \midrule
    \multirow{2}[2]{*}{\shortstack{ParSeD\\(VG, OD)}} & \cellcolor{lightgray} 1\%   & \cellcolor{lightgray} 60    & \cellcolor{lightgray} 0.18 & \cellcolor{lightgray} 2.05 & \cellcolor{lightgray} 1.62 \\
          & \cellcolor{darkgray} 10\%  & \cellcolor{darkgray} 60 & \cellcolor{darkgray} 6.46 & \cellcolor{darkgray} 12.19 & \cellcolor{darkgray} 10.87 \\
    \midrule
    \multirow{2}[2]{*}{\shortstack{ParSeD\\(VG, RD)}} & \cellcolor{lightgray} 1\%   & \cellcolor{lightgray} 60    & \cellcolor{lightgray} 3.74 & \cellcolor{lightgray} 8.62 & \cellcolor{lightgray} 7.50 \\
          & \cellcolor{darkgray} 10\%  & \cellcolor{darkgray} 60    & \cellcolor{darkgray} 12.29 & \cellcolor{darkgray} 17.98 & \cellcolor{darkgray} 16.67 \\
    \midrule
    \multirow{2}[2]{*}{\shortstack{ParSeD\\(COCO, OD)}} & \cellcolor{lightgray} 1\%   & \cellcolor{lightgray} 60    & \cellcolor{lightgray} 5.86  & \cellcolor{lightgray} 10.16  & \cellcolor{lightgray} 9.17  \\
          & \cellcolor{darkgray} 10\%  & \cellcolor{darkgray} 60    & \cellcolor{darkgray} 12.20  & \cellcolor{darkgray} 20.39  & \cellcolor{darkgray} 18.51  \\
    \midrule
    \multirow{3}[2]{*}{\shortstack{RLIP-ParSeD\\(VG)}} & 0\%   & -     & 12.30  & 12.81  & 12.69  \\
          & \cellcolor{lightgray} 1\%   & \cellcolor{lightgray} 10    & \cellcolor{lightgray} 16.24  & \cellcolor{lightgray} 16.05  & \cellcolor{lightgray} 16.09  \\
          & \cellcolor{darkgray} 10\%  & \cellcolor{darkgray} 10    & \cellcolor{darkgray} 15.43  & \cellcolor{darkgray} 20.34  & \cellcolor{darkgray} 19.21  \\
    \midrule
    \multirow{3}[2]{*}{\shortstack{RLIP-ParSeD\\(COCO + VG)}} & 0\%   & -     & 11.20  & 14.73  & 13.92  \\
           & \cellcolor{lightgray} 1\%   & \cellcolor{lightgray} 10    & \cellcolor{lightgray} 16.22  & \cellcolor{lightgray} 18.92 & \cellcolor{lightgray} 18.30  \\
          & \cellcolor{darkgray} 10\%  & \cellcolor{darkgray} 10 & \cellcolor{darkgray} 15.89 & \cellcolor{darkgray} 23.94 & \cellcolor{darkgray} 22.09  \\
    \bottomrule
    \end{tabular}%
    \label{few_shot_transfer_RLIP_ParSeD}%
\end{minipage}
\vspace{-0.3cm} 
\end{minipage}

\paragraph{The influence of relation label noise.}
To assess sensitivity to noise, we report fine-tuning results on HICO-DET with increasing ratios of relation label noise in~\cref{RLIP_Label_Noise}.
We observe that as label noise increases, the COCO object detection pre-training adopted by prior work exhibits a greater degradation in performance (29.12$\rightarrow$24.52, -4.60) than RLIP (29.21$\rightarrow$25.68, -3.53)
We also observe that when initialising RLIP-ParSeD with COCO pre-trained parameters, RLIP again helps to ameliorate noise, with a more limited loss of performance (30.70$\rightarrow$26.87, -3.83) than COCO pre-training and a similar degradation to RLIP-ParSeD with random initialization.
Consequently, we deduce that RLIP offers a route to mitigating label corruption and improving model robustness~\cite{hendrycks2019pretrain_robustness}.

\begin{minipage}[!t]{\textwidth}
\begin{minipage}[c]{0.35\textwidth}
    \small
    \setlength{\tabcolsep}{2pt}
    \centering
        \includegraphics[width=0.7\textwidth]{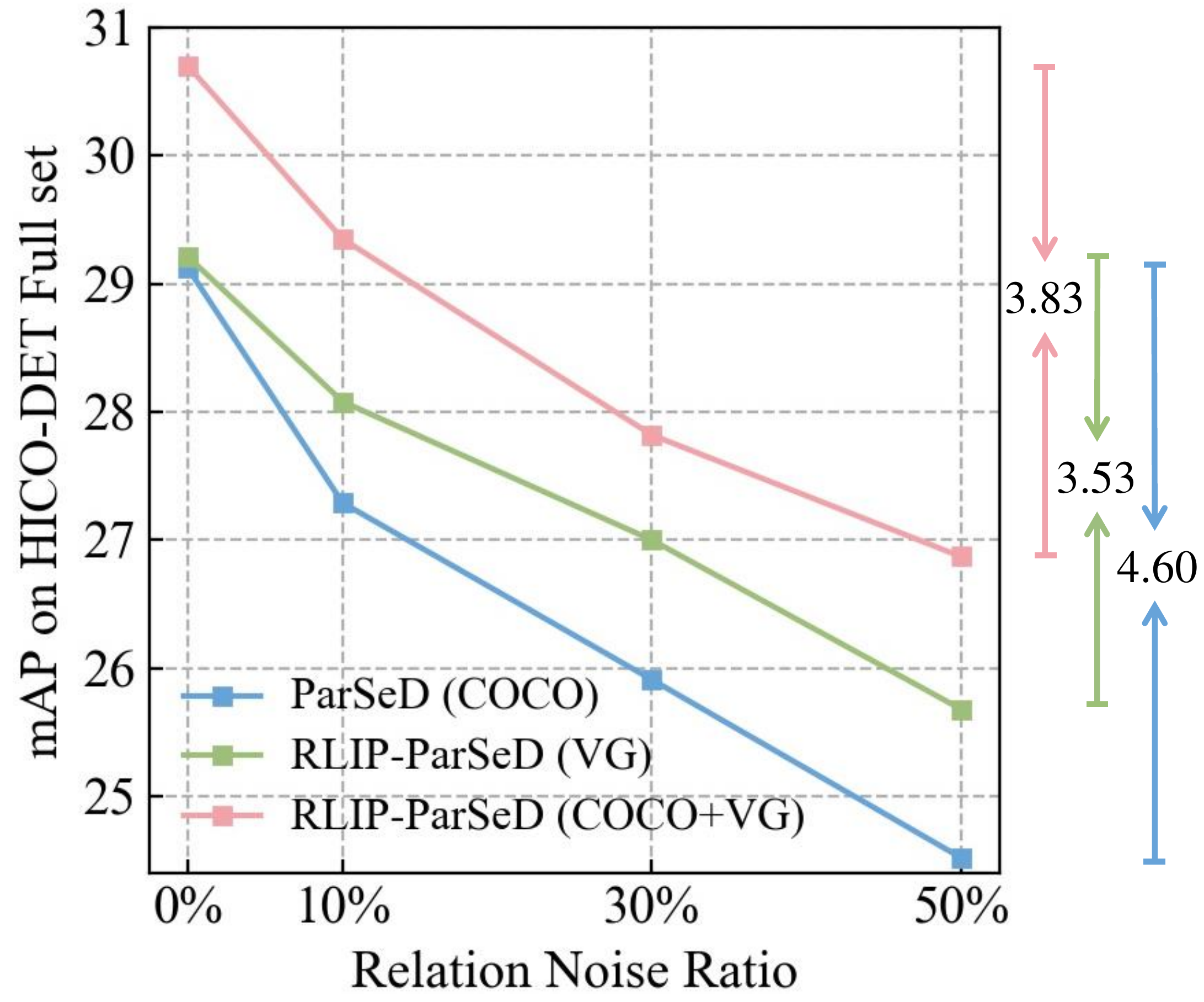}
        \makeatletter\def\@captype{figure}\makeatother\caption{Relation Label Noise.}
        \label{RLIP_Label_Noise}
\end{minipage}
\begin{minipage}[c]{0.6\textwidth}
    \centering
    \small
    \setlength{\tabcolsep}{2pt}
    \centering
    \makeatletter\def\@captype{table}\makeatother\caption{Ablation study of different sampling strategies for label sequence extension using RLIP-ParSeD on HICO-DET.}
    \begin{tabular}{ccccccc}
        \toprule
        \multirow{2}[2]{*}{\textbf{Sampling Type}} & \multicolumn{3}{c}{\textbf{Fine-tuning}} & \multicolumn{3}{c}{\textbf{Zero-shot (NF)}} \\
              & \textbf{Rare} & \textbf{Non-Rare} & \textbf{Full} & \textbf{Rare} & \textbf{Non-Rare} & \textbf{Full} \\
        \midrule
        -     & 22.58  & 28.98  & 27.51  & 9.77  & 9.97  & 9.92  \\
        Uniform & 23.33  & 29.55  & 28.12  & 9.46  & 9.67  & 9.63  \\
        Frequency-based & 23.02  & 29.77  & 28.22  & 10.45  & 11.26  & 11.07  \\
        \bottomrule
    \end{tabular}%
  \label{ablation_sampling_strategy}%
\end{minipage}
\vspace{-0.3cm} 
\end{minipage}

\subsection{Ablation studies and analysis}
\vspace{-0.2cm} 

\paragraph{Ablation study of ParSe on the influence of decoupled representations.}
We report an ablation study of the ParSe architecture in~\cref{ablation_ParSe_main_paper} to highlight the importance of decoupling the representation of subjects, objects and relations.
The first row of~\cref{ablation_ParSe_main_paper} represents the use of coupled representations for subjects, objects and relations~\cite{tamura2021qpic}.
The second row of~\cref{ablation_ParSe_main_paper} represents the use of coupled representations for subjects and objects that are disentangled from relations~\cite{zhang2021CDN}.
The final row (ParSe) uses fully-disentangled representations.
We observe a clear gain resulting from ParSe over methods using a joint representation of (some subset of) subject, object and relation triplets.

\paragraph{Ablation study of sampling strategy.}
In~\cref{ablation_sampling_strategy}, we present an ablation study to assess the efficacy of out-of-batch sampling strategies.
Intuitively, \textit{uniform sampling} will up-weight descriptions from the tail of the distribution while \textit{frequency-based sampling} will preserve the distribution.
Both bring improvements to fine-tuning by providing additional negative samples.
However, by over-sampling descriptions from the tail, uniform sampling performs less well with common texts in the downstream task, and thus fares less well overall.
%

\begin{wrapfigure}{r}{0.35\textwidth}
  \vspace{-0.6cm}
  \begin{center}
    \includegraphics[width=0.35\textwidth]{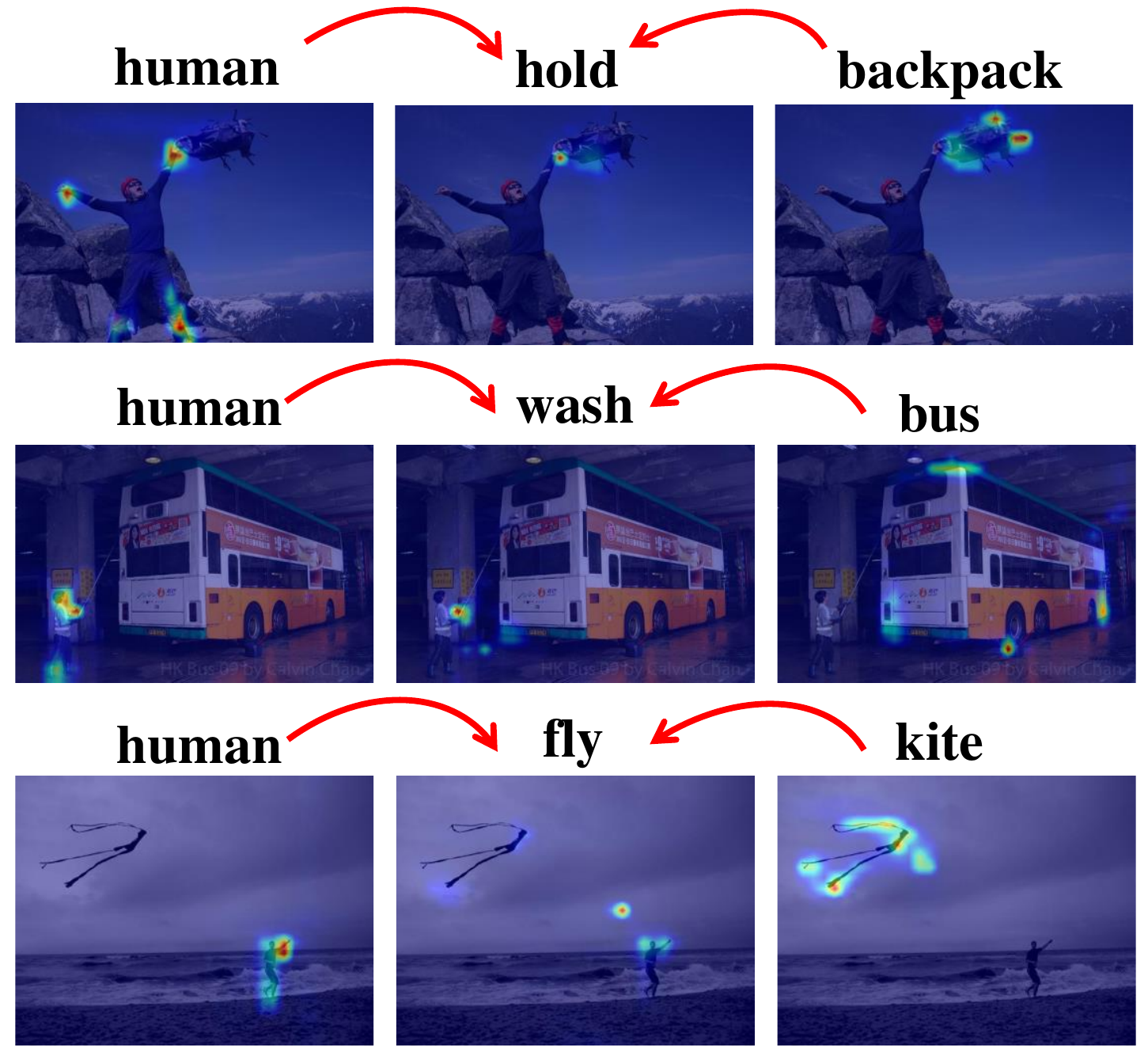}
  \end{center}
  \caption{\small Attention weight analysis for the top-1 scored verb. Weights are extracted from \textit{Parallel Entity Detection} (human and object) and \textit{Sequential Relation Inference} (verb).}
  \vspace{-0.8cm}
  \label{attn_weights_vis}
\end{wrapfigure}

\textbf{Visualisation of \ParSe attention weights.}
We visualise attention weights from \ParSe for several example images in~\cref{attn_weights_vis}.
We observe that \ParSe attends distinct regions for subjects, objects and relations.
This aligns with our motivation that entity detection is best supported by local context, while relation inference draws on additional spatial context, conditioning on subjects and objects like hands and string, as well as the wet ground and sky where appropriate.
%

\textbf{Ablation study of \rlip techniques.}
In~\cref{ablation_RLIP_technique}, we present an ablation study of the three proposed technical contributions.
We observe that each benefits both fine-tuning and zero-shot NF under all metrics.
LSE attains a greater boost for the \textit{Non-Rare} set (by sampling according to the training set distribution).
On the other hand, RPL enhances results for the \textit{Rare} set, likely due to the propensity of  RPL to label rare descriptions as positive (due to the long-tailed distribution of text labels).
%

\begin{table}[t]
  \centering
  \caption{Fine-tuning results with ParSe (COCO, OD) on HICO-DET.}
    \begin{tabular}{llccc}
    \toprule
    \textbf{ParSe Architecture} & \textbf{Coupling} &\textbf{Rare} & \textbf{Non-Rare} & \textbf{Full} \\
    \midrule
    - & coupled subject, objects and relations    & 23.18  & 31.45  & 29.55  \\
    w/ Se & coupled subject and objects & 25.58  & 32.50  & 30.91  \\
    w/ ParSe & fully decoupled & \textbf{26.36}  & \textbf{33.41}  & \textbf{31.79}  \\
    \bottomrule
    \end{tabular}
    \vspace{-0.4cm} 
  \label{ablation_ParSe_main_paper}
\end{table}

\begin{table}[t]
  \small
  \setlength{\tabcolsep}{4pt}
  \centering
  \caption{Ablation study of RLIP techniques using RLIP-ParSeD on HICO-DET.}
    \begin{tabular}{ccccccccccc}
    \toprule
    \multicolumn{3}{c}{\textbf{RLIP Technique}} & \multicolumn{3}{c}{\textbf{Fine-tuning}} & \multicolumn{3}{c}{\textbf{Zero-shot (NF)}} & \multicolumn{2}{c}{\textbf{Relation}} \\
    \textbf{LSE} & \textbf{RQL} & \textbf{RPL} & \textbf{Rare} & \textbf{Non-Rare} & \textbf{Full} & \textbf{Rare} & \textbf{Non-Rare} & \textbf{Full} & \textbf{Uniformity$\downarrow$} & \textbf{Alignment$\downarrow$} \\
    \midrule
          &       &       & 22.58  & 28.98  & 27.51  & 9.77  & 9.97  & 9.92 & -0.8233 & 0.3650 \\
    \checkmark     &       &       & 23.02  & 29.77  & 28.22  & 10.45  & 11.26  & 11.07 & -1.0556 & 0.4542 \\
    \checkmark     & \checkmark     &       & 24.32  & 30.32  & 28.94  & 11.49  & 12.60  & 12.34 & -1.3986 & 0.6072 \\
    \checkmark     & \checkmark     & \checkmark     & 24.45  & 30.63  & 29.21  & 12.30  & 12.81  & 12.69 & -1.3265 & 0.5799 \\
    \bottomrule
    \end{tabular}%
    \vspace{-0.4cm} 
  \label{ablation_RLIP_technique}%
\end{table}%

\textbf{Understanding LSE, RQL and RPL with Uniformity and Alignment.}
To gain insight into representation quality, Wang et al.~\cite{wang2020uniformity} proposed two metrics,  \textit{uniformity} and \textit{alignment}, which we employ here to better understand the influence of our contributions.
To this end, we perform a zero-shot (NF) evaluation on HICO-DET, using bipartite matching to assign predicted triplets to ground-truth labels.
We then calculate uniformity and alignment metrics for relation features via
$\mathcal{L}_{u}(f;t) = {\rm log}(\mathbb{E}_{(x,y) \stackrel {\rm i.i.d}\sim \mathcal{P}_{data}} [e^{-t\|f(x) - f(y) \|_2^{2}}])$ and $\mathcal{L}_{a}(f;\alpha) = \mathbb{E}_{(x,y) \sim \mathcal{P}_{pos}} [\|f(x) - f(y) \|_2^{\alpha}]$, where $\alpha, t = 2$.
We present the results in~\cref{ablation_RLIP_technique}, where we observe that LSE and RQL both reduce uniformity, the former through additional negative pairs and the latter by reducing the loss assigned to over-confident text labels. 
We also observe that RPL yields better alignment through discovering additional positive labels.
The results indicate, however, a trade-off between the metrics---a useful direction for future work would be to determine how to find an appropriate balance between them.

\paragraph{Verb-wise mAP Analysis for zero-shot (NF) evaluation.}
We provide analysis to give a sense of the verb overlap of HICO with VG.
We use ``relationship aliases'' from the official VG website to obtain as many HOI verb annotations from VG as possible by string matching.
The result is shown in \cref{tab:verb_overlap_vg_hico} in the supplementary material. 
We observe that in VG there are only 2,203 HOI verb annotations even when considering relationship aliases---approximately 1.47\% of the number of relationship annotations in HICO-DET.
30 HOI verbs do not have an annotation and 45 HOI verbs have five or fewer annotations. 
In RLIP-ParSe (COCO+VG), we observe that mAP for the 30 verbs is 5.56 while mAP for the remaining 87 verbs is 18.12.
If we use uni-modal relation detection pre-training, the result for the 30 verbs degrades to zero.
In light of this, we conjecture that existing relations can transfer their knowledge to the inference of non-existing relations in HOI detection.
To provide a more detailed analysis, we show the verb-wise mAP on HICO verbs in VG (\cref{fig:mAP_HICO_verb_in_VG}) and not in VG (\cref{fig:mAP_HICO_verb_not_in_VG}) with zero-shot (NF) evaluation (figures are provided in the supplementary material), where we observe solid performance for some verbs.

\paragraph{Probing into reasons for the verb zero-shot performance.}
We aim to \textbf{\textit{qualitatively understand where the zero-shot ability stems from}}.
In the above analysis, \textit{pay} has the highest performance among verbs not seen by VG (\cref{fig:mAP_HICO_verb_not_in_VG} in the supplementary material).
In the methodology section, we present the conditional query generation that constrains the verb inference to be related to subjects and objects, providing verb inference with a conditional context.
Thus, to analyze how this ability of verb zero-shot inference emerges, we need to consider the subject and object context as they are essential to predict the verb in ParSe.
For the verb \textit{pay} in HICO-DET, there is only one possible triplet annotated, "person pay parking meter". Then, we want to answer, "\textbf{Is there any triplet annotated with similar or identical subjects and objects that transfer the inference ability to \textit{pay}?}"
To answer this question, we search for triplets annotated with similar subjects and objects to HICO-DET from VG (For details, please refer to the analysis of \cref{tab:verb_dist_similar_triplets} in the supplementary material.). 
We report the verb distribution of the limited number of triplets that are found, ranking the verbs in ascending order of Euclidean distance to the target verb (\cref{tab:verb_dist_similar_triplets_one_example}). 
From this table, we can see that the verbs quantitatively closer (in Euclidean distance or Cosine distance) to \textit{pay} have similar meanings to \textit{pay}, shown by their lexical variants or grammatical variants (e.g., \textit{putting money in} has a similar meaning to \textit{pay}).
Thus, in the VG dataset, there is \textit{human putting money in parking meter}, which may transfer to the zero-shot recognition of \textit{person pay parking meter} in HICO-DET.
More examples can be found in \cref{tab:verb_dist_similar_triplets}.
In short conclusion, we conjecture that the zero-shot inference ability of RLIP is not from the scale of annotations (by comparing relation detection pre-training and RLIP using VG), but the ability to transfer the verb inference knowledge from semantically similar annotations. This analysis also accords with previous works~\cite{radford2021CLIP,li2021GLIP} that semantic diversity is important as it introduces large-scale potential annotations, ensuring a model transfers well to different data distributions.

\begin{table}[t]
    \vspace{-0.2cm}
    \centering
    \small
    \setlength{\tabcolsep}{2pt}
    \caption{VG verb ranking given similar subject-object triplets from HICO-DET. Verbs are in ascending order of Euclidean distance. (The Cosine distance can also output similar rankings.)}
    \begin{tabular}{cccccccc}
        \toprule
        \shortstack{\textbf{"pay"}\\("parking meter")} & \textbf{putting money in} & \textbf{collecting money at} & \textbf{puts change into} & \textbf{repairing} & \textbf{checking} & \textbf{next to} & ... \\
        \midrule
        Count & 1 & 1 & 1 & 1 & 1 & 1 & ... \\
        Euclidean & 11.56 & 11.70 & 13.34 & 14.21 & 15.16 & 16.12 & ... \\
        Cosine & 0.4560 & 0.4576 & 0.3108 & 0.2554 & 0.1583 & 0.0709 & ... \\
        \bottomrule
    \end{tabular}
    \vspace{-0.6cm}
    \label{tab:verb_dist_similar_triplets_one_example}
\end{table}


\begin{wraptable}{r}{0.5\textwidth}
    \vspace{-0.5cm}
    \centering
    \small
    \setlength{\tabcolsep}{4pt}
    \caption{Uniformity analysis of the seen verbs, unseen verbs and all verbs before and after RLIP. Lower uniformity value is better.}
    \begin{tabular}{cccc}
        \toprule
        \textbf{Verb Set} & \textbf{Seen (87)} & \textbf{Unseen (30)} & \textbf{All (117)} \\
        \midrule
        Before RLIP & -0.00367 & -0.00436 & -0.00388 \\
        After RLIP & -3.73780 & -3.59457 & -3.71330 \\
        \bottomrule
    \end{tabular}
    \label{tab:verb_uniformity_main_paper}
    \vspace{-0.3cm}
\end{wraptable}

Second, we aim to \textbf{\textit{demonstrate quantitatively how RLIP pre-trains the model to perform zero-shot detection from the perspective of representation learning}}. 
We employ the Uniformity metric introduced in~\cite{wang2020uniformity}. 
Uniformity is a metric to assess a model's generalization in contrastive learning. 
In this case, since label textual embeddings serve as a classifier in RLIP, we calculate the Uniformity of the seen verbs, unseen verbs and all verbs, aiming to observe how the generalization changes before and after RLIP, and how the generalization varies between seen verbs and unseen verbs. The results are shown in \cref{tab:verb_uniformity_main_paper}.
As can be seen from the table, Uniformity values are high before RLIP, suggesting that the representations before RLIP are distributed compactly, leading to a poor classifier.
However, after RLIP is performed, the 87 seen verbs have a substantially lower Uniformity value, corresponding with decent zero-shot performance. 
Similarly, the 30 unseen verbs and the combination of 117 verbs also have excellent Uniformity values, contributing to unseen zero-shot performance. 
Through this quantitative observation, we think that from the perspective of representations, RLIP contributes to improved zero-shot performance.

From all the above analysis, we conjecture that the zero-shot performance is not caused by the increased dataset size or annotations, but rather from the generalization in representations obtained by pre-training with language supervision.


%% file: 05-conclusion.tex
\vspace{-0.2cm} 
\section{Conclusion} \label{sec:conclusion}
\vspace{-0.2cm} 
In this paper, we propose RLIP as a pre-training strategy for HOI detection.
We show that RLIP, together with our additional technical contributions, boosts HOI detection performance under fine-tuning, zero-shot and few-shot evaluations, and improves robustness against noisy annotations.

\paragraph{Acknowledgements:}
We would like to appreciate anonymous reviewers for their valuable feedback and members from Fundamental Vision Intelligence Team of Alibaba DAMO Academy for sharing computational resources. This work was supported in part by National Natural Science Foundation of China Grant No. 62173298 and by Alibaba Group through Alibaba Research Intern Program.



%% file: 08-appendix.tex
\appendix
\section{Appendix}
In this supplementary material, we first discuss the potential societal impact (\cref{app:impact}) and limitations (\cref{app:limitations}) of our approach.
Next, we provide further details on the architecture of \rlipParSeD (\cref{app:arch}), dataset pre-processing (\cref{app:preproc}), phased pre-training (\cref{app:phased-pretraining}), attention analysis (\cref{app:attention}) and subject-object query pairing (\cref{app:sub-obj-pairing}).
Finally, we provide additional experiments and analysis (\cref{app:additional-exps}) and discuss our use of datasets (\cref{app:datasets}). Codes will be publicly available upon publication.

\subsection{Potential Societal Impact} \label{app:impact}

By targeting improved HOI detection, our work has the potential to bring societal and commercial benefits in medical, retail, security and sports analysis applications. 
However, it is inherently a dual-use technology, providing functionality with scope for abuse.
For example, better HOI detection may make it easier to conduct unlawful surveillance.
Moreover, due to biases present among training datasets, it is also likely that our system does not perform equally across all demographics.
Therefore, we caution that our approach represents a research proof-of-concept and is not suitable for real-world usage without a rigorous evaluation of the deployment context and appropriate oversight.

\subsection{Limitations and Potential Future Works} \label{app:limitations}

As noted in the experiment section of the main paper, one limitation of our method is its dependence on a particular form of annotations (\textit{i.e.} the pre-training data must be annotated with relation triplets), which are not always available, or exist at a diminished scale relative to object detection annotations.
In our work, we investigated one potential solution by bootstrapping RLIP from object detection parameters to mitigate annotation scarcity.
Although existing relation annotations are limited, we do not anticipate that this will remain the case. Indeed, we hope that our work will inspire future work to focus on this problem and dataset contributions will follow. 
Besides, we provide ways to scale up datasets as future works.
For example, we could reuse a grounding dataset with entities annotated. Then, a language processing tool like spaCy~\cite{honnibal2017spacy} can be adopted as a tool to obtain their relations from captions. Even if we do not have subjects and objects but only image-caption pairs, we can combine the use of spaCy and methods like GLIP~\cite{li2021GLIP} to create abundant triplet annotations. Based on the analysis, we think our method is still promising and inspiring, paving a path for further research.

A second limitation of our approach is its requirement for long optimisation schedules during pre-training---a characteristic that is inherited from DETR~\cite{carion2020DETR}.
Although DDETR~\cite{zhu2020deformableDETR} ameliorates this issue to some extent, it achieves weaker performance. 
Thus, one potential for future work is to combine the benefits of DETR's high performance and DDETR's fast convergence speed.

\subsection{Overall Architecture of \rlipParSeD} \label{app:arch}
\begin{figure}[t]
\centering
\includegraphics[width=1\textwidth]{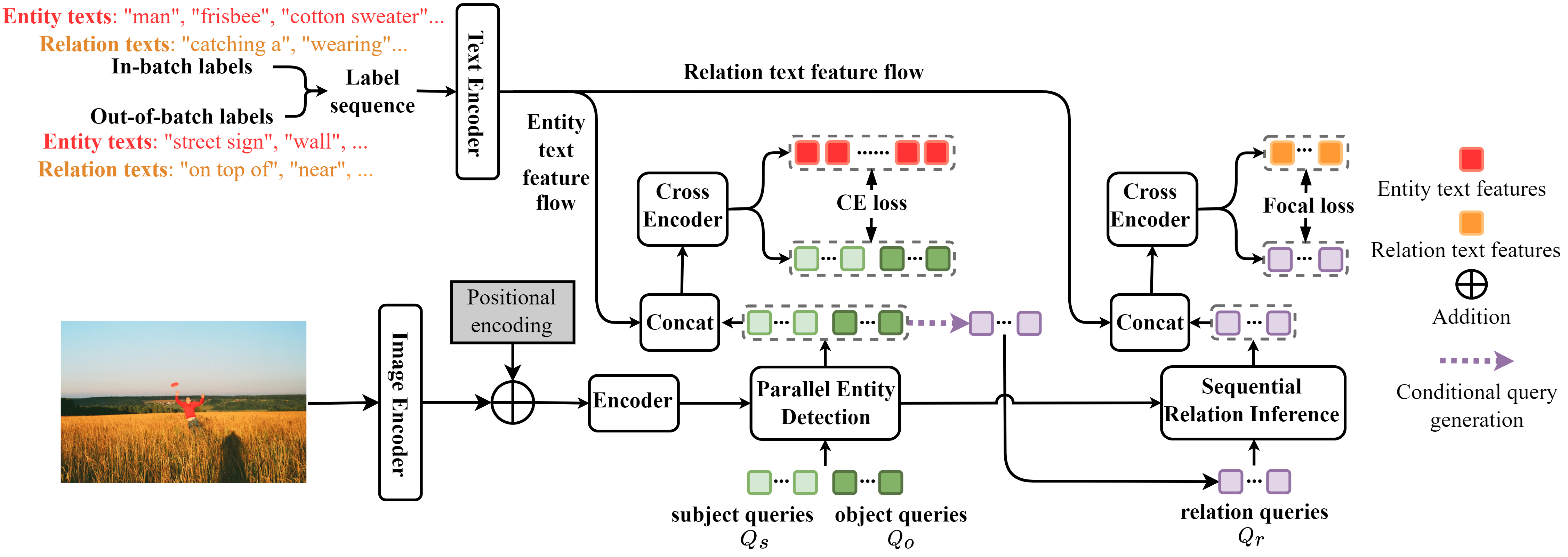}
\caption{\small An overview of our pre-training framework, \rlipParSeD.
The encoder represents DDETR-style encoders.
The \textit{Parallel Entity Detection} and \textit{Sequential Relation Inference} blocks represent independent DDETR-style decoders responsible for entity and relation detection, respectively.
The cross encoder represents DETR-style encoders.
We omit the localisation loss for clarity.}
\vspace{-0.4cm}
\label{RLIP_ParSeD_pipeline}
\end{figure}

We present the detailed architecture of \rlipParSeD as shown in~\cref{RLIP_ParSeD_pipeline}.
The major difference between \rlipParSe (shown in the main paper) and \rlipParSeD is the cross-modal fusion module.
For the former, we use the detection Transformer encoder to directly fuse language-image features as previous works have done~\cite{li2021ALBEF,kamath2021MDETR,akbari2021VATT}. 
For the latter, we use an additional Transformer encoder to fuse the decoded queries and language features following~\cite{Maaz2021MViT} since deformable attention~\cite{zhu2020deformableDETR} from DDETR relies on spatial coordinates, which language features do not have.
Note that the decoded subject and object queries are fused with entity text features and the decoded relation queries are fused with relation text features.
The two cross encoders do not share parameters.
The localisation loss is applied on decoded queries (after \textit{Parallel Entity Detection} and before the cross encoder).
Other architecture/\rlip details follow the descriptions provided in the main paper.

\subsection{Pre-Processing Steps for Visual Genome} \label{app:preproc}
Due to the crowd-sourcing process used to construct Visual Genome~\cite{krishna2017visualgenome}, there are many redundant annotations.
Thus, we conduct basic cleaning steps to filter out such annotations and all of our experiments are conducted on the dataset after pre-processing.
The steps are listed as follows:
\begin{itemize}
    \item We keep the first object text description for every object because a very small proportion of objects have multiple text descriptions.
    \item We filter out redundant triplets by \textbf{i)} keeping only one if there are multiple identical triplets and \textbf{ii)} keeping only one if there are multiple triplets with identical subject descriptions, object descriptions, relation descriptions and similar box locations (with both the subject box's and the object box's IoU>0.5)
    \item We filter out redundant triplets if the number of triplets in one image is greater than the number of queries $N_{Q} = 100$.
\end{itemize}

\subsection{Additional Details for Phased Pre-training} \label{app:phased-pretraining}
In the experiments section of the main paper, we describe \textbf{i)} how object detection parameters (obtained via pre-training on COCO) can be used to initialise \rlipParSe and \rlipParSeD and \textbf{ii)} how MDETR~\cite{kamath2021MDETR} parameters (obtained via pre-training on GoldG+) can be used to initialise MDETR-ParSe.
Here, we provide further details on how these are implemented. 

For the main blocks of \rlipParSe pre-trained on COCO, we initialise the image encoder, cross encoder, \textit{Parallel Entity Detection} block and \textit{Sequential Relation Inference} block parameters with parameters from the image encoder, detection Transformer encoder, detection Transformer decoder (first 3 layers) and detection Transformer decoder (first 3 layers) of a COCO pre-trained DETR model following \cite{zhang2021CDN}. 
We initialise the FFN layers of \rlipParSe pre-trained on COCO using the localization FFN layer parameters and entity classification FFN layer parameters (since object categories in HICO-DET and V-COCO are identical to COCO).
For \rlipParSeD pre-trained on COCO, parameter initialisation follows \rlipParSe.
Other parameters are randomly initialised.

For the main blocks of MDETR-ParSe~\cite{kamath2021MDETR}, we initialise the parameters of the text encoder, image encoder, cross encoder, \textit{Parallel Entity Detection} block and \textit{Sequential Relation Inference} block with parameters from the text encoder, image encoder, cross encoder, detection Transformer decoder (first 3 layers) and detection Transformer decoder (first 3 layers) of a GoldG+ pre-trained MDETR model.
For the FFN layers of MDETR-ParSe, we initialise from localization FFN layers.
Other parameters are randomly initialised.


\subsection{Details of Attention Weight Analysis} \label{app:attention}
In Figure 3 of the main paper, we provide an analysis of the attention weights produced by ParSe (extracted from ParSe in RLIP-ParSe). Here, we provide additional details of how this analysis is conducted.
During image inference, we employ a Transformer~\cite{AttentionAlluNeed} architecture to decode queries. 
In this decoding process, a ${\rm softmax}$ function is used to normalize attention weights calculated by a scaled dot-product attention.
The logits after the ${\rm softmax}$ function indicate the importance of regions (since queries aggregate values according to the logits).
Thus, we extract the logits after the ${\rm softmax}$ function in the last Transformer layer of both the \textit{Parallel Entity Detection} and the \textit{Sequential Relation Inference} block for the top-1 scored verb.
To visualize the attention weights, we linearly scale the range of logits to 0--255 (and cast to integers to produce an image).

\subsection{Computational Overhead of the Subject and Object Query Pairing} \label{app:sub-obj-pairing}
The pairing of humans and objects are performed by index-matching as is stated in the main paper. Thus, we pair humans and objects with identical indices (e.g., the first decoded feature from the subject queries and the first decoded feature from the object queries are paired.). Due to the simplicity of this matching strategy, the cost is trivial ($\mathcal{O}(1)$ cost) compared to the overall overhead during model inference.

\subsection{Additional Experiments and Analysis} \label{app:additional-exps}

\paragraph{Ablation study of ParSe on the influence of decoupled representations.}
We report a further ablation study of the ParSe architecture in~\cref{ablation_ParSe} to highlight the importance of decoupling the representation of subjects, objects and relations.
The first row of~\cref{ablation_ParSe} represents the use of coupled representations for subjects, objects and relations~\cite{tamura2021qpic}.
The second row of~\cref{ablation_ParSe} represents the use of coupled representations for subjects and objects that are disentangled from relations~\cite{zhang2021CDN}.
The final row (ParSe) uses fully-disentangled representations.
We observe a clear gain resulting from ParSe over methods using a joint representation of (some subset of) subject, object and relation triplets.

\begin{table}[h]
  \centering
  \caption{Fine-tuning results with ParSe (COCO) on HICO-DET.}
    \begin{tabular}{llccc}
    \toprule
    \textbf{ParSe Architecture} & \textbf{Coupling} &\textbf{Rare} & \textbf{Non-Rare} & \textbf{Full} \\
    \midrule
    - & coupled subject, objects and relations    & 23.18  & 31.45  & 29.55  \\
    w/ Se & coupled subject and objects & 25.58  & 32.50  & 30.91  \\
    w/ ParSe & fully decoupled & \textbf{26.36}  & \textbf{33.41}  & \textbf{31.79}  \\
    \bottomrule
    \end{tabular}
  \label{ablation_ParSe}
\end{table}%

\paragraph{Robustness towards different backbones.} Compare more thoroughly with CDN~\cite{zhang2021CDN} and QAHOI~\cite{chen2021qahoi}, we perform extensive experiments to demonstrate the effectiveness of the uni-modal detection pipeline ParSe as shwon in~\cref{results_diff_backbones}. As the table indicates, ParSe outperforms CDN-L with half the number of decoding layers and a single-stage fine-tuning with a clear gain (+0.69 mAP on Full set). When compared to QAHOI, ParSe improves by 1.18 mAP on Full set with only two fifths number of fine-tuning epochs. If using the same number of epochs, ParSe can surpass it by 1.97 mAP on Full set and more improvement on the Rare set (+3.32mAP).

\begin{table}[h]
    \centering
    \small
    \setlength{\tabcolsep}{4pt}
    \caption{Fully-finetuned results on HICO-DET with different backbones. PTP, DL and PT denote Pre-training paradigm, decoding layers and pre-training.}
    \begin{tabular}{ccccccccc}
        \toprule
        \textbf{Method} & \textbf{Backbone} & \textbf{DL} & \textbf{PTP} & \textbf{PT data} & \textbf{\#Tuning Epochs} & \textbf{Rare} & \textbf{Non-Rare} & \textbf{Full} \\ 
        \midrule
        CDN-L~\cite{zhang2021CDN} & ResNet-101 & 12 & OD & COCO & 90+10 & 27.19 & 33.53 & 32.07 \\
        ParSe & ResNet-101 & 6 & OD & COCO & 90 & 28.59 & 34.01 & 32.76 \\
        \midrule
        QAHOI~\cite{chen2021qahoi} & Swin-T & 6 & - & - & 150 & 22.44 & 30.27 & 28.47 \\
        ParSe & Swin-T & 6 & - & - & 60 & 23.77 & 31.40 &29.65 \\
        ParSe & Swin-T & 6 & - & - & 150 & 25.76 & 31.84 & 30.44 \\
        \bottomrule
    \end{tabular}
    \label{results_diff_backbones}
\end{table}

\paragraph{Superiority over other models with VG and COCO data} To have a fairer comparison with previous methods using VG and COCO data, we adopt CDN~\cite{zhang2021CDN} as a base method and then add the VG dataset to its pre-training stage. Since RLIP also adopts the relation annotations in VG, we also try to include these annotations in the uni-modal pre-training. Thus, we resort to relation detection on VG. To be more specific, we perform uni-modal relation detection pre-training by using linear classifiers for verbs and entities rather than matching with texts. The results are shown in \cref{tab:vg_coco_CDN_ParSe}. We can see from the table that by using uni-modal relation detection pre-training, CDN still trails RLIP-ParSe with the same number of epochs of pre-training and fine-tuning, which shows the effectiveness of RLIP. Even if comparing it with ParSe using relation detection pre-training, we can still observe an improvement of ParSe over CDN, demonstrating the usefulness of decoupling triplet representations.

\begin{table}[h]
    \centering
    \small
    \setlength{\tabcolsep}{4pt}
    \caption{Fully-finetuned results on HICO-DET with VG and COCO dataset. RD and PT denote relation detection and pre-training.}
    \begin{tabular}{cccccccc}
        \toprule
        \textbf{Method} & \textbf{Detector} & \textbf{Data} & \textbf{PT Paradigm} & \textbf{PT \#Epochs} & \textbf{Rare} & \textbf{Non-Rare} & \textbf{Full} \\ 
        \midrule
        CDN & DETR & COCO+VG & Relation Detction & 150 & 25.65 & 32.75 & 31.12 \\
        ParSe & DETR & COCO+VG & Relation Detction & 150 & 26.00 & 33.40 & 31.70 \\
        RLIP-ParSe & DETR & COCO+VG & RLIP & 150 & 26.85 & 34.63 & 32.84 \\
        \bottomrule
    \end{tabular}
    \label{tab:vg_coco_CDN_ParSe}
\end{table}

\paragraph{Few-shot transfer with \rlipParSe.}
To evaluate few-show transfer with ParSe and \rlipParSe, we fine-tune ParSe for 90 epochs as above, while \rlipParSe is fine-tuned for 10 epochs to avoid over-fitting.
The detailed results are shown in~\cref{tab:few_shot_transfer_RLIP_ParSe} and evaluation curves are shown in~\cref{fig:few_shot_transfer_RLIP_ParSe}.
We observe that \rlip with object detection initialisation significantly benefits few-shot fine-tuning relative to object detection pre-training in terms of final results and convergence speed, especially when data is scarce.
\rlipParSe with 1\% data achieves similar performance with ParSe with 10\% data.

\begin{minipage}[!t]{\textwidth}
\begin{minipage}[c]{0.45\textwidth}
    \small
    \setlength{\tabcolsep}{2pt}
    \centering
        \includegraphics[width=0.8\textwidth]{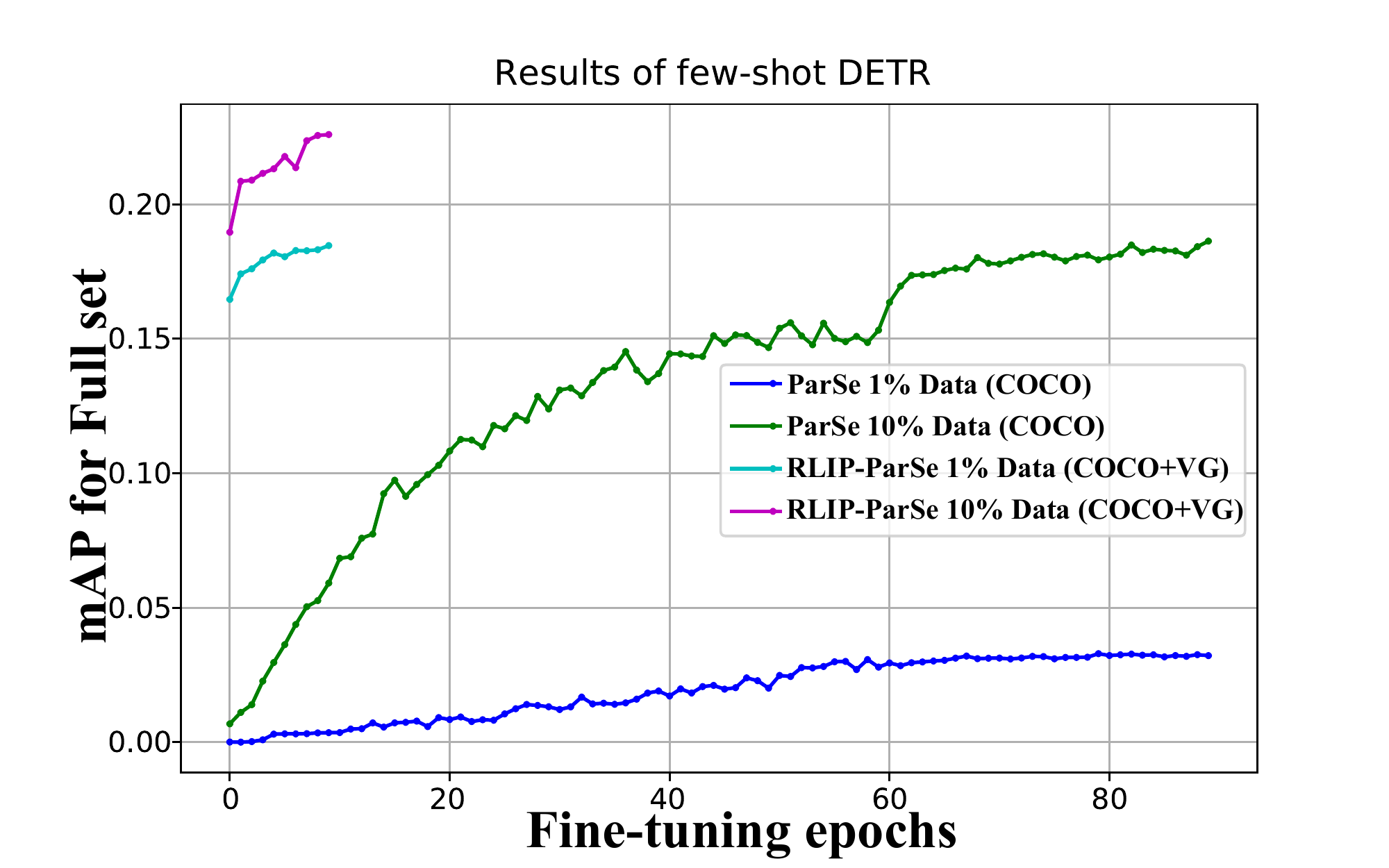}
        \makeatletter\def\@captype{figure}\makeatother\caption{\small Evaluation curves for few-shot transfer on HICO-DET with ParSe and \rlipParSe.}
        \label{fig:few_shot_transfer_RLIP_ParSe}
\end{minipage}
\begin{minipage}[c]{0.55\textwidth}
    \small
    \setlength{\tabcolsep}{3pt}
    \centering
    \makeatletter\def\@captype{table}\makeatother\caption{Few-shot transfer on HICO-DET with ParSe and \rlipParSe.}
    \begin{tabular}{cccccc}
    \toprule
    \textbf{Method} & \textbf{Data} & \textbf{Epochs} & \textbf{Rare} & \textbf{Non-Rare} & \textbf{Full} \\
    \midrule
    \multirow{2}[2]{*}{\shortstack{ParSe\\(COCO)}} & \cellcolor{lightgray} 1\%  & \cellcolor{lightgray} 90 & \cellcolor{lightgray} 1.69  & \cellcolor{lightgray} 3.67  & \cellcolor{lightgray} 3.21 \\
          & \cellcolor{darkgray} 10\%  & \cellcolor{darkgray} 90 & \cellcolor{darkgray} 14.61 & \cellcolor{darkgray} 19.56 & \cellcolor{darkgray} 18.42 \\
    \midrule
    \multirow{3}[2]{*}{\shortstack{RLIP-ParSe\\(COCO+VG)}} & 0\%   & - & 15.08 & 15.50 & 15.40 \\
            & \cellcolor{lightgray} 1\%   & \cellcolor{lightgray} 10 & \cellcolor{lightgray} 17.47 & \cellcolor{lightgray} 18.76 & \cellcolor{lightgray} 18.46 \\
            & \cellcolor{darkgray} 10\%  & \cellcolor{darkgray} 10 & \cellcolor{darkgray} 20.16 & \cellcolor{darkgray} 23.32 & \cellcolor{darkgray} 22.59 \\
    \bottomrule
    \end{tabular}%
    \label{tab:few_shot_transfer_RLIP_ParSe}%
\end{minipage}
\end{minipage}

\paragraph{Detailed results of relation label noise with \rlipParSe and \rlipParSeD} We present detailed results concerning the influence of relation label noise on RLIP-ParSeD and ParSeD in~\cref{tab:relation_label_noise_RLIP_ParSeD} (which corresponds to Figure 2 in the main paper) and the influence of relation label noise on RLIP-ParSe and ParSe in~\cref{tab:relation_label_noise_RLIP_ParSe}.
The RLIP-ParSe and ParSe results support our claim that RLIP helps to ameliorate noise since ParSe suffers a greater degradation of performance (31.79$\rightarrow$25.19, -6.6) than RLIP-ParSe (32.84$\rightarrow$27.75, -5.09).

\begin{minipage}[!t]{\textwidth}
\begin{minipage}[c]{0.5\textwidth}
    \centering
    \small
    \setlength{\tabcolsep}{2pt}
    \centering
    \makeatletter\def\@captype{table}\makeatother\caption{\small Relation label noise on HICO-DET with ParSeD and \rlipParSeD.}
    \begin{tabular}{ccccc}
    \toprule
    \textbf{Method} & \textbf{Noise} & \textbf{Rare} & \textbf{Non-Rare} & \textbf{Full} \\
    \midrule
    \multirow{4}[2]{*}{\shortstack{ParSeD\\(COCO)}} & 0\%   & 22.23  & 31.17  & 29.12  \\
          & 10\%  & 19.63  & 29.58  & 27.29  \\
          & 30\%  & 17.14  & 28.52  & 25.91  \\
          & 50\%  & 15.82  & 27.12  & 24.52 \\
    \midrule
    \multirow{4}[2]{*}{\shortstack{RLIP-ParSeD\\(VG)}} & 0\%   & 24.45  & 30.63  & 29.21  \\
          & 10\%  & 21.59  & 30.02  & 28.08  \\
          & 30\%  & 19.60  & 29.21  & 27.00  \\
          & 50\%  & 17.11  & 28.24  & 25.68 \\
    \midrule
    \multirow{4}[2]{*}{\shortstack{RLIP-ParSeD\\(COCO+VG)}} & 0\%   & 24.67  & 32.50  & 30.70  \\
          & 10\%  & 19.86  & 32.20  & 29.35  \\
          & 30\%  & 18.45  & 30.62  & 27.82  \\
          & 50\%  & 17.81  & 29.58  & 26.87 \\
    \bottomrule
    \end{tabular}%
    \label{tab:relation_label_noise_RLIP_ParSeD}%
\end{minipage}
\begin{minipage}[c]{0.5\textwidth}
  \setlength{\tabcolsep}{2pt}
  \small
  \centering
    \makeatletter\def\@captype{table}\makeatother\caption{\small Relation label noise on HICO-DET with ParSe and \rlipParSe.}
    \begin{tabular}{ccccc}
    \toprule
    \textbf{Method} & \textbf{Noise} & \textbf{Rare} & \textbf{Non-Rare} & \textbf{Full} \\
    \midrule
    \multicolumn{1}{c}{\multirow{4}[2]{*}{\shortstack{ParSe\\(COCO)} }} & 0\%   & 26.36  & 33.41  & 31.79  \\
          & 10\%  & 21.59  & 30.80  & 28.68  \\
          & 30\%  & 20.52  & 29.99  & 27.81  \\
          & 50\%  & 15.01  & 28.23  & 25.19  \\
    \midrule
    \multicolumn{1}{c}{\multirow{4}[2]{*}{\shortstack{RLIP-ParSe\\(COCO+VG)} }} & 0\%   & 26.85  & 34.63  & 32.84  \\
          & 10\%  & 24.62  & 33.54  & 31.49  \\
          & 30\%  & 23.12  & 31.75  & 29.77  \\
          & 50\%  & 20.09  & 30.04  & 27.75  \\
    \bottomrule
    \end{tabular}%
  \label{tab:relation_label_noise_RLIP_ParSe}%
\end{minipage}
\end{minipage}

\paragraph{Sensitivity analysis of hyper-parameter $\eta$ in RPL.}
In~\cref{varying_eta}, we present a sensitivity analysis for the $\eta$ hyperparamter used in RPL.
As $\eta$ decreases, RPL selects more descriptions with high similarities as positive, boosting performance.
However, if $\eta$ is too low, there is an increased risk of false positives arising in the pseudolabeling process.
We choose $\eta=0.3$ in the main paper according to this experiment.

\begin{table}[h]
  \centering
  \caption{Results with varying $\eta$ in RPL with \rlipParSeD on HICO-DET. LSE and RQL are used by default.}
    \begin{tabular}{ccccccc}
    \toprule
    \multirow{2}[2]{*}{$\eta$} & \multicolumn{3}{c}{\textbf{Fine-tuning}} & \multicolumn{3}{c}{\textbf{Zero-shot (NF)}} \\
          & \textbf{Rare} & \textbf{Non-Rare} & \textbf{Full} & \textbf{Rare} & \textbf{Non-Rare} & \textbf{Full} \\
    \midrule
    0.2   & 24.05  & \textbf{30.73} & 29.19  & 10.95  & 12.71  & 12.30  \\
    0.3   & \textbf{24.45} & 30.63  & \textbf{29.21} & \textbf{12.30} & \textbf{12.81} & \textbf{12.69} \\
    0.4   & 23.67  & 29.90  & 28.47  & 11.97  & 12.80  & 12.61  \\
    0.5   & 22.63  & 29.79  & 28.14  & 11.70  & 12.09  & 12.00  \\
    \bottomrule
    \end{tabular}
  \label{varying_eta}
\end{table}%

\paragraph{Design choice of distance function $m(\cdot ,\cdot)$ in RPL.} For the design choice of $m(\cdot ,\cdot)$, we also experiment on another widely-adopted distance function Cosine distance. We conduct a sensitivity analysis of the hyper-parameter $\eta$ to compare with the one chosen in the main paper (last row of results). The zero-shot (NF) results of Cosine distance using RLIP-ParSeD is shown in \cref{Cosine_distance}. We observe that Euclidean distance is slightly better (the last row of results is selected in the paper). Since both methods have similar computational overhead, in the paper, we choose the Euclidean distance.

\begin{table}[h]
    \centering
    \caption{Zero-shot (NF) results with varying $\eta$ in RPL with \rlipParSeD on HICO-DET. LSE and RQL are used by default.}
    \begin{tabular}{cccccc}
        \toprule
        \textbf{Distance Function} & $\bm{\eta}$ & \textbf{Rare} & \textbf{Non-Rare} & \textbf{Full} \\ 
        \midrule
        Cosine & 0.3 & 11.21 & 12.53 & 12.23 \\ 
        Cosine & 0.4 & 11.92 & 12.82 & 12.61 \\ 
        Cosine & 0.5 & 11.76 & 12.71 & 12.49 \\ 
        Cosine & 0.6 & 11.30 & 12.22 & 12.01 \\ 
        \textbf{Euclidean} & \textbf{0.3} & \textbf{12.30} & \textbf{12.81} & \textbf{12.69} \\
        \bottomrule
    \end{tabular}
    \label{Cosine_distance}%
\end{table}

\paragraph{Comparing RQL and RPL with Label Smoothing Regularization.}
Since we employ RPL and RQL to smooth the target distributions used during training to account for ambiguity, it is useful to compare their effectiveness to a manually-designed Label Smoothing Regularisation (LSR) method, such as the one introduced in~\cite{szegedy2016inceptionv3}.
We experiment with using LSR for relation labels (following RPL and RQL).
However, we find that LSR tends to degrade performance, while the proposed RPL and RQL approaches boost all metrics.

\begin{table}[h]
  \centering
  \caption{Results comparing Label Smoothing Regularization (LSR)~\cite{szegedy2016inceptionv3} with RPL+RQL on HICO-DET. We use RLIP-ParSeD with LSE as a base model.}
    \begin{tabular}{ccccccc}
    \toprule
    \multicolumn{1}{c}{\multirow{2}[2]{*}{\textbf{\shortstack{Ambiguity\\suppression}}}} & \multicolumn{3}{c}{\textbf{Fine-tuning}} & \multicolumn{3}{c}{\textbf{Zero-shot (NF)}} \\
          & \textbf{Rare} & \textbf{Non-Rare} & \textbf{Full} & \textbf{Rare} & \textbf{Non-Rare} & \textbf{Full} \\
    \midrule
    -     & 23.02  & 29.77  & 28.22  & 10.45  & 11.26  & 11.07  \\
    LSR   & 23.51  & 29.38  & 28.03  & 10.03  & 10.84  & 10.65  \\
    RPL+RQL & \textbf{24.45}  & \textbf{30.63}  & \textbf{29.21}  & \textbf{12.30}  & \textbf{12.81}  & \textbf{12.69}  \\
    \bottomrule
    \end{tabular}%
  \label{Compare_RQL_RPL_with_LSR}%
\end{table}%

\paragraph{Similarity analysis between in-batch labels and out-of-batch labels} The in-batch labels are aggregated from images' annotations, and the out-of-batch labels are sampled from the whole dataset, which does not overlap with in-batch labels. Since the contrastive loss optimizes to push away the negative textual labels, we can observe the change of the similarities of the negative and positive labels. To quantitatively analyze it, we simulate the training process by out-of-batch sampling, and observe the change of similarities by calculating the average pairwise distance of the positive labels to the negative labels. We mainly compare the object and relation similarity based on the RoBERTa model before and after RLIP pre-training. The results are shown in~\cref{label_similarity_analysis}. From this table, we can see that the Cosine similarity decreases, and Euclidean distance increases. Note that before RLIP, the discrimination ability of text embeddings are poor, which corresponds with previous work~\cite{representation_degrdation_NLP}. The results indicate whichever distance function we adopt (Cosine or Euclidean distance) and whichever kind of feature we observe (object or relation), the similarity between in-batch labels and out-of-batch labels decreases after performing RLIP. This enables the language model to adapt well to the visual representations and serve as a good classifier.

\begin{table}[h]
    \centering
    \small
    \setlength{\tabcolsep}{5pt}
    \caption{Similarity analysis between in-batch labels and out-of-batch labels before and after RLIP. Cos and Euc abbreviate Cosine distance and Euclidean distance.}
    \begin{tabular}{ccccc}
    \hline
        \toprule
        \textbf{Model} & \textbf{Object (Cos)} & \textbf{Relation (Cos)} & \textbf{Object (Euc)} & \textbf{Relation (Euc)} \\
        \midrule
        \textbf{RoBERTa (Before RLIP)} & 0.9991 & 0.9986 & 0.2502 & 0.3156 \\
        \textbf{RoBERTa (After RLIP)} & 0.0084 & 0.0208 & 18.1943 & 16.9177 \\
        \bottomrule
    \end{tabular}
    \label{label_similarity_analysis}%
\end{table}

\paragraph{Verb-wise mAP analysis for zero-shot (NF) evaluation} We provide analysis to give a sense of the verb overlap of HICO with VG. We use ``relationship aliases'' from the official VG website to obtain as many HOI verb annotations from VG as possible. The result is shown in \cref{tab:verb_overlap_vg_hico}. 

\begin{table}[h]
    \centering
    \caption{Verb overlap of HICO with VG.}
    \begin{tabular}{ccccc}
        \toprule
        \textbf{Dataset} & \textbf{\#images} & \textbf{HOI verb annos} & \textbf{HOI verb annos' ratio} & \textbf{Imbalance ratio} \\
        \midrule
        VG & 108K & 2,203 & 1.47\% & 304 \\
        \bottomrule
    \end{tabular}
    \label{tab:verb_overlap_vg_hico}
\end{table}

We observe that in VG, there are 2,203 HOI verb annotations even when considering relationship aliases---approximately 1.47\% of the number of relationship annotations in HICO-DET. 30 HOI verbs do not have an annotation and 45 HOI verbs have five or fewer annotations. 
In RLIP-ParSe (COCO+VG), we observe that mAP for the 30 verbs is 5.56 while mAP for the remaining 87 verbs is 18.12. If we use uni-modal relation detection pre-training, the result for the 30 verbs degrades to zero.

\begin{figure}[h]
    \centering
    \includegraphics[width=1\textwidth]{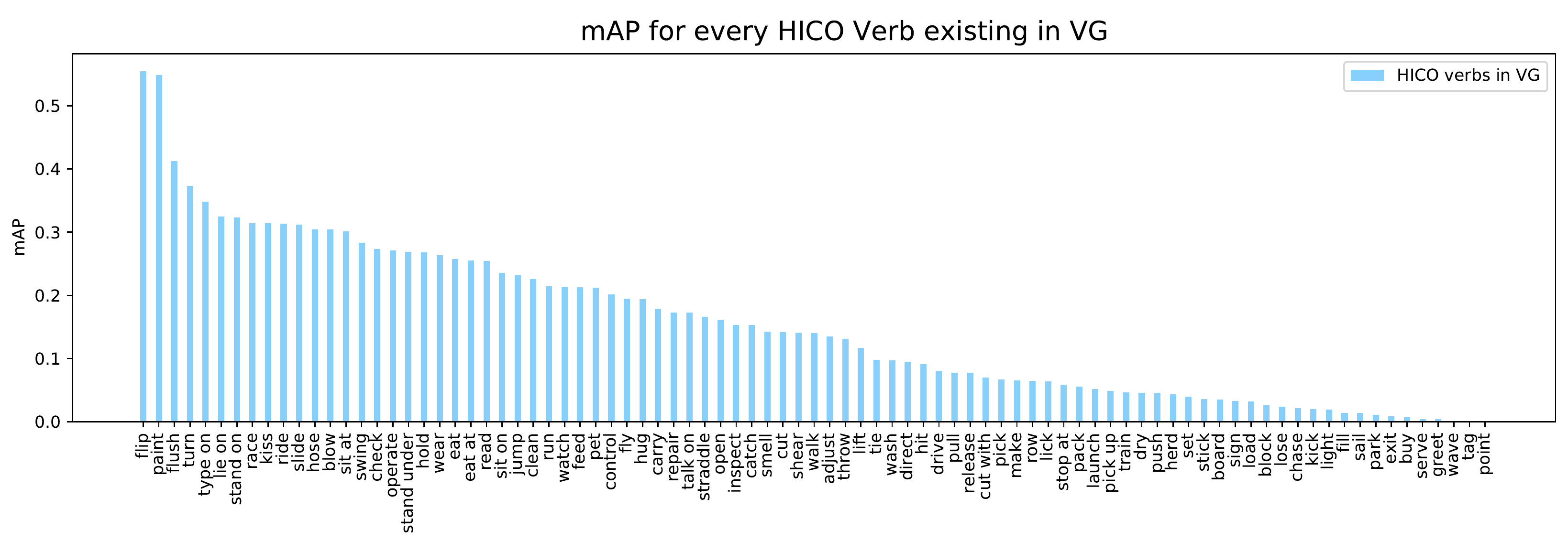}
    \caption{\small Verb-wise mAP Analysis for Zero-Shot (NF) Evaluation. Presented verbs exist in VG.}
    \vspace{-0.4cm}
    \label{fig:mAP_HICO_verb_in_VG}
\end{figure}

\begin{figure}[h]
    \centering
    \includegraphics[width=0.4\textwidth]{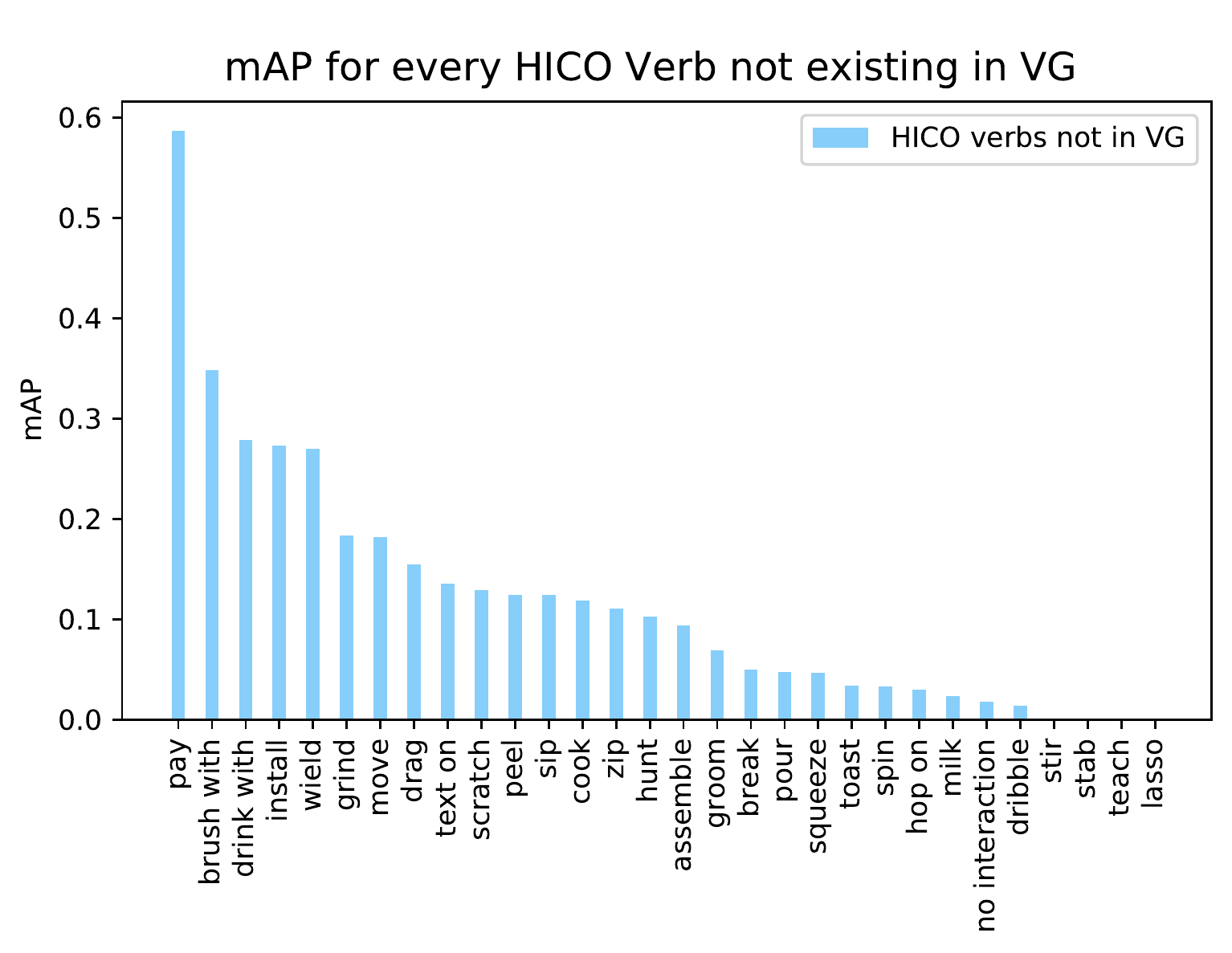}
    \caption{\small Verb-wise mAP Analysis for Zero-Shot (NF) Evaluation. Presented verbs do not exist in VG.}
    \vspace{-0.4cm}
    \label{fig:mAP_HICO_verb_not_in_VG}
\end{figure}

To provide a more detailed analysis, we show the verb-wise mAP on HICO verbs in VG (\cref{fig:mAP_HICO_verb_in_VG}) and not in VG (\cref{fig:mAP_HICO_verb_not_in_VG}) with zero-shot (NF) evaluation. We can observe solid performance for some verbs.

\paragraph{Probing into reasons for the verb zero-shot performance} \label{paragraph:probe_into_zero-shot}
We aim to \textbf{\textit{qualitatively understand where the zero-shot ability stems from}}.
In the above analysis, \textit{pay} has the highest performance among verbs not seen by VG. 
In the main paper, we present the conditional query generation that constrains the verb inference to be related to subjects and objects, providing verb inference with a conditional context. 
Thus, to analyze how this ability of verb zero-shot inference emerges, we need to consider the subject and object context as they are essential to predict the verb in ParSe. 
For the verb \textit{pay} in HICO-DET, there is only one possible triplet annotated, "person pay parking meter". Then, we want to answer, "\textbf{is there any triplet annotated with similar or identical subjects and objects that transfer the inference ability to \textit{pay}?}"
To answer this question, we search for triplets annotated with similar subjects and objects to HICO-DET from VG. For the subjects, we heuristically select ones whose textual descriptions have any one of the following strings: \textit{man, woman, person, friend, guy, dude, human, people, driver, passenger, hand, limb}. For the objects, we heuristically select ones whose textual descriptions have the string of the target object. 
By this processing, only a limited number of triplets are found. 
Building on these, we report the verb distribution of the limited number of triplets that are found, ranking the verbs in ascending order of Euclidean distance to the target verb, results of which are shown in \cref{tab:verb_dist_similar_triplets}. 
From this table, we can see that the verbs quantitatively closer (in Euclidean distance or Cosine distance) to \textit{pay} have similar meanings to \textit{pay}, shown by their lexical variants or grammatical variants (e.g., \textit{putting money in} has a similar meaning to \textit{pay}).
Thus, in the VG dataset, there is \textit{human putting money in parking meter}, which may transfer to the zero-shot recognition of \textit{person pay parking meter} in HICO-DET.
Similarly, we could see that in the VG datatset, there is \textit{human putting condiments on hot dog}, which may transfer to the zero-shot recognition of \textit{person cook hot dog} in HICO-DET.
Note that there are some grammatical variations that we can not exhaustively set rules to avoid, thus creating a zero-shot setting which is not theoretically strict enough.
But we want the model to benefit from this property of natural language in the context of language-image pre-training.

In short conclusion, with the assistance of the sequential inference structure of verbs, we think that the zero-shot inference ability in RLIP is not from the scale of annotations (by comparing relation detection pre-training and RLIP using VG), but the ability to transfer the verb inference knowledge from semantically similar annotations. This analysis also accords with previous papers~\cite{radford2021CLIP,li2021GLIP} that semantic diversity is important as it introduces large-scale potential annotations, ensuring a model transfers well to different data distributions.

\begin{table}[h]
    \vspace{-0.2cm}
    \centering
    \small
    \setlength{\tabcolsep}{2pt}
    \caption{VG verb ranking given similar subject-object triplets from HICO-DET. Verbs are in ascending order of Euclidean distance. (The Cosine distance can also output similar rankings.)}
    
    \begin{tabular}{cccccccc}
        \toprule
        \shortstack{\textbf{"pay"}\\("parking meter")} & \textbf{putting money in} & \textbf{collecting money at} & \textbf{puts change into} & \textbf{repairing} & \textbf{checking} & \textbf{next to} & ... \\
        \midrule
        Count & 1 & 1 & 1 & 1 & 1 & 1 & ... \\
        Euclidean & 11.56 & 11.70 & 13.34 & 14.21 & 15.16 & 16.12 & ... \\
        Cosine & 0.4560 & 0.4576 & 0.3108 & 0.2554 & 0.1583 & 0.0709 & ... \\
        \bottomrule
    \end{tabular}
    
    \begin{tabular}{cccccccc}
        \toprule
        \shortstack{\textbf{"cook"}\\("hot dog")} & \textbf{putting condiments on} & \textbf{prepping} & \textbf{displeased with} & \textbf{roasts} & \textbf{blowing on} & \textbf{about to eat} & ... \\
        \midrule
        \textbf{Count} & 1 & 1 & 1 & 1 & 1 & 1 & ... \\
        \textbf{Euclidean} & 13.38 & 14.14 & 15.48 & 15.63 & 16.04 & 16.27 & ... \\
        \textbf{Cosine} & 0.3467 & 0.2565 & 0.0656 & 0.1787 & 0.0471 & 0.0680 & ... \\
        \bottomrule
    \end{tabular}
    
    \begin{tabular}{cccccccc}
        \toprule
        \shortstack{\textbf{"grind"}\\("skateboard")} & \textbf{race downhill} & \textbf{flying off ramp on} & \textbf{skating} & \textbf{midair on} & \textbf{for balancing on} & \textbf{competes on} & ... \\
        \midrule
        \textbf{Count} & 1 & 1 & 2 &  & 1 & 1 &... \\
        \textbf{Euclidean} & 13.26 & 13.27 & 13.44 & 13.59 & 14.26 & 14.36 & ... \\
        \textbf{Cosine} & 0.3670 & 0.3401 & 0.3553 & 0.3288 & 0.2510 & 0.2037 &... \\
        \bottomrule
    \end{tabular}
    
    \begin{tabular}{cccccccc}
        \toprule
        \shortstack{\textbf{"assemble"}\\("kite")} & \textbf{are preparing their} & \textbf{launched} & \textbf{managing} & \textbf{launch} & \textbf{carry} & \textbf{directing} & ... \\
        \midrule
        \textbf{Count} & 1 & 1 & 2 & 1 & 1 & 2 & ... \\
        \textbf{Euclidean} & 12.48 & 12.72 & 13.13 & 13.20 & 13.31 & 13.71 & ... \\
        \textbf{Cosine} & 0.3350 & 0.3684 & 0.2974 & 0.2554 & 0.2027 & 0.2565 & ... \\
        \bottomrule
    \end{tabular}
    
    \begin{tabular}{cccccccc}
        \toprule
        \shortstack{\textbf{"text on"}\\("cell phone")} & \textbf{viewing messages on} & \textbf{typing on} & \textbf{texting} & \textbf{speaking on} & \textbf{listening on} & \textbf{speaks on} & ... \\
        \midrule
        \textbf{Count} & 1 & 1 & 2 & 1 & 1 & 2 & ... \\
        \textbf{Euclidean} & 10.24 & 11.24 & 11.42 & 11.85 & 11.85 & 11.98 & ... \\
        \textbf{Cosine} & 0.5876 & 0.4437 & 0.3983 & 0.4082 & 0.4281 & 0.3557 & ... \\
        \bottomrule
    \end{tabular}
    
    \begin{tabular}{cccccccc}
        \toprule
        \shortstack{\textbf{"scratch"}\\("dog")} & \textbf{touching} & \textbf{bent over touching} & \textbf{touches} & \textbf{interacting with} & \textbf{holding a hot} & \textbf{reaching for} & ... \\
        \midrule
        \textbf{Count} & 8 & 1 & 2 & 1 & 1 & 2 & ... \\
        \textbf{Euclidean} & 14.02 & 14.47 & 14.54 & 14.60 & 14.80 & 14.81 & ... \\
        \textbf{Cosine} & 0.2157 & 0.1566 & 0.2260 & 0.0969 & 0.0512 & 0.2039 & ... \\
        \bottomrule
    \end{tabular}
    
    \vspace{-0.2cm}
    \label{tab:verb_dist_similar_triplets}
\end{table}

Secondly, we aim to \textbf{\textit{demonstrate quantitatively how RLIP pre-trains the model to perform zero-shot detection from the perspective of representation learning}}. 
We resort to the Uniformity metric introduced in~\cite{wang2020uniformity}. 
Uniformity is a metric to assess a model's generalization in contrastive learning. We detail the calculation of this metric in the analysis of the main paper. 
In this case, since label textual embeddings serve as a classifier in RLIP, we calculate the Uniformity of the seen verbs, unseen verbs and all verbs, aiming to observe how the generalization changes before and after RLIP, and how the generalization varies between seen verbs and unseen verbs. The results are shown in \cref{tab:verb_uniformity}.
As can be seen from the table, Uniformity values are all high before RLIP. It means that the representations before RLIP are compactly distributed, serving as an awful classifier. 
However, after RLIP is performed, the seen 87 verbs have a distinctively lower Uniformity value, corresponding with the decent zero-shot performance. 
Similarly, the 30 unseen verbs and the combination of 117 verbs also have excellent Uniformity values, contributing to the unseen zero-shot performance. 
Through this quantitative observation, we think that from the perspective of representations, RLIP contributes to the real zero-shotness.

From all the above analysis, we think that the zero-shotness may not be caused by the mounting dataset size or annotations, but stem from the generalization in representations obtained by pre-training with language supervision.

\begin{table}[h]
    \vspace{-0.3cm}
    \centering
    \small
    \setlength{\tabcolsep}{4pt}
    \caption{Uniformity analysis of the seen verbs, unseen verbs and all verbs before and after RLIP. Lower uniformity value is better.}
    \begin{tabular}{cccc}
        \toprule
        \textbf{Verb Set} & \textbf{Seen (87)} & \textbf{Unseen (30)} & \textbf{All (117)} \\
        \midrule
        Before RLIP & -0.00367 & -0.00436 & -0.00388 \\
        After RLIP & -3.73780 & -3.59457 & -3.71330 \\
        \bottomrule
    \end{tabular}
    \label{tab:verb_uniformity}
    \vspace{-0.3cm}
\end{table}

\paragraph{The influence of semantic-diverse data (upstream data distributions)} We ablate semantic diversity by significantly altering the distribution of VG annotations and assessing the influence on RLIP's performance. To this end, first note that since VG is human-annotated with free-form text, it is extremely long-tailed. We alter its distribution by dropping tail object classes and verb classes to create a dataset with limited semantic diversity. Concretely, we drop object classes whose instance counts are fewer than 1,000 and relation classes whose instance counts are fewer than 500. We pre-train RLIP on the resulting dataset and then perform zero-shot (NF) evaluation on HICO-DET. The results are shown in \cref{tab:influence_semantic_diversity}. We observe from this table that despite a very significant change to the training distribution, performance on the Full set drops only moderately. We do, however, witness a relatively larger decline on the Rare set due to the lack of semantic diversity in the modified data. This finding accords with the observations of GLIP~\cite{li2021GLIP}. To make full use of language-image pre-training, semantic diversity is important which can ensure a good domain transfer as is indicated by CLIP~\cite{radford2021CLIP} and GLIP~\cite{li2021GLIP}.

\begin{table}[h]
    \vspace{-0.3cm}
    \centering
    \small
    \setlength{\tabcolsep}{3pt}
    \caption{Semantic diversity analysis with zero-shot (NF) evaluation on HICO-DET. Obj, rel and annos denote object, relation and annotations respectively.}
    \begin{tabular}{ccccccccc}
        \toprule
        \textbf{Method} & \textbf{Data} & \textbf{Obj classes} & \textbf{Obj annos} & \textbf{Rel classes} & \textbf{Rel annos} & \textbf{Rare} & \textbf{Non-Rare} & \textbf{Full} \\
        \midrule
        ParSeD & VG & 100,298 & 3.80m & 36,515 & 1.99m & 12.30 & 12.81 & 12.69 \\
        ParSeD & VG- & 497 & 1.73m & 151 & 1.27m & 9.45 & 12.13 & 11.51 \\
        \bottomrule
    \end{tabular}
    \vspace{-0.3cm}
    \label{tab:influence_semantic_diversity}
\end{table}

\begin{figure}[h]
    \centering
    \includegraphics[width=0.8\textwidth]{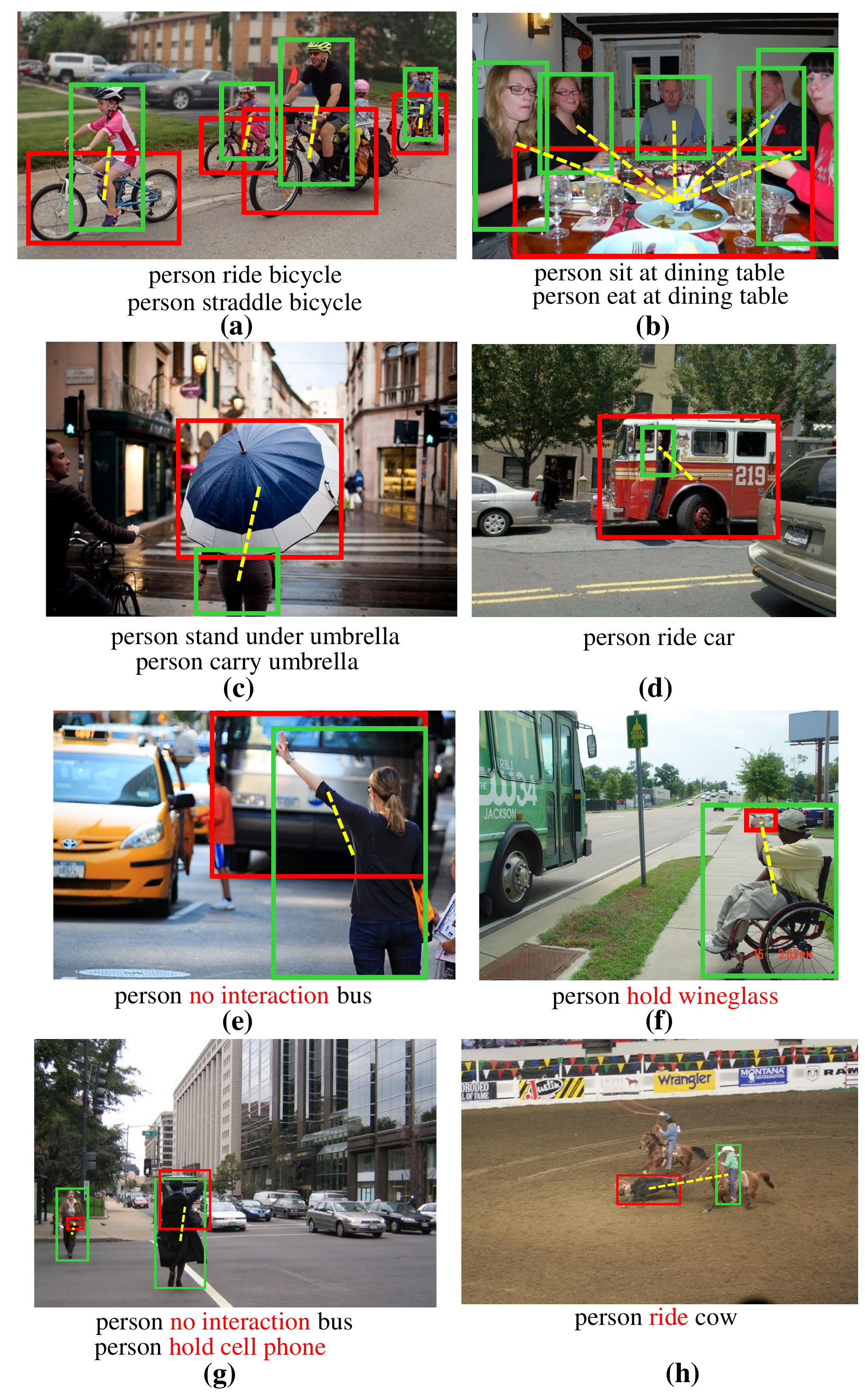}
    \caption{\small Successful and failure case analysis. We visualize the true positives (a-d) with verb scores higher than 0.3, and
    the false positives (e-h) from top-3 predictions produced by the RLIP-ParSeD (VG) model. Words in \textcolor{red}{red} indicate wrong predictions.}
    \vspace{-0.4cm}
    \label{fig:failure_cases}
\end{figure}

\paragraph{More successful and failure cases analysis and corresponding potential future work} In this work, we present ParSe as an effective HOI detection structure. In \cref{fig:failure_cases} (a-d), we present several cases where the model successfully predicts by using RLIP-ParSeD (VG), with verb scores greater than 0.3. From \cref{fig:failure_cases} (a) and (b), we can see that although the scene is complex with many possible triplets with multiple labels, the model can detect the right subjects and objects, linking them as triplets. From \cref{fig:failure_cases} (c) and (d), we can observe that although the person is only partially visible, the model still detects him/her and then predicts the right triplets. Both cases show that RLIP-ParSeD can overcome difficult cases during application.
However, there are also failure cases, where further works can improve upon. 
We show the failure cases from top-3 predictions produced by RLIP-ParSeD (VG) in \cref{fig:failure_cases} (e-h).
\textbf{i)} First of all, the verb inference in ParSe conditions on the detection results since Sequential Relation Inference is fed with queries generated by detection features. 
Also, VG does not provide a good object detection foundation.
Thus, an inferior detection result can lead to false positives.
As shown in \cref{fig:failure_cases}(f) and (g), the model detects the objects to be \textit{wine glass} and \textit{cell phone} (which should be \textit{fire hydrant} and \textit{handbag}), thus producing wrong predictions.
Building on this observation, we think excellent object detectors can be designed and incorporated into HOI detection.
\textbf{ii)} Secondly, DETR-based models may find it hard to detect fine-grained poses in people, which degrades the performance of inferring some verbs like \textit{wave}. As shown in \cref{fig:failure_cases} (e) and (g), the model fails to predict \textit{wave} although right localizing the subject person. Building on this observation, we think it to be promising to efficiently incorporate pose cues in end-to-end HOI detection models.
\textbf{iii)} Thirdly, some contextual cues may be hard to be detected. As shown in \cref{fig:failure_cases} (h), the model predicts \textit{ride} rather than \textit{lasso}. If the model captures the rope as a global context, then it can perform well. Building on this observation, we think it to be promising to efficiently incorporate fine-grained contextual cues or introduce external object knowledge to detect the triplet.

\subsection{Datasets used in this work} \label{app:datasets}
\textbf{Licenses.}
The V-COCO~\cite{gupta2015VisualSemanticRole} dataset is used under an MIT license.
The HICO-DET~\cite{chao2015hico,chao2018learningtodetectHOI} dataset is used under a CC0: Public Domain license.
The Visual Genome~\cite{krishna2017visualgenome} dataset is used under a Creative Commons Attribution 4.0 International License.
\\
\textbf{Release of personally identifiable information/offensive content/consent.} 
We do not release data as part of this research.
We work with standard public domain benchmarks for computer vision: Visual Genome~\cite{krishna2017visualgenome},
HICO-DET~\cite{chao2015hico,chao2018learningtodetectHOI},
and V-COCO~\cite{gupta2015VisualSemanticRole}.
We therefore assess that the risk of releasing personally identifiable information or offensive content is relatively low. 
With regards to consent, we did not pursue an independent investigation of consent that goes beyond the considerations of the original dataset releases.